\documentclass{article}

\PassOptionsToPackage{numbers, compress}{natbib}
\usepackage[preprint]{neurips_2024}

\usepackage[utf8]{inputenc} % allow utf-8 input
\usepackage[T1]{fontenc}    % use 8-bit T1 fonts
\usepackage{hyperref}       % hyperlinks
\usepackage{url}            % simple URL typesetting
\usepackage{booktabs}       % professional-quality tables
\usepackage{amsfonts}       % blackboard math symbols
\usepackage{nicefrac}       % compact symbols for 1/2, etc.
\usepackage{microtype}      % microtypography
\usepackage{xcolor}         % colors
\usepackage{graphicx}
\usepackage{amsmath}
\usepackage{cleveref}
\usepackage{enumitem}
\usepackage{wrapfig}
\graphicspath{{res/}}

\newcommand{\Tau}{\mathcal{T}}

\title{Cell Morphology-Guided Small Molecule Generation with GFlowNets}

\author{%
  Stephen Zhewen Lu \\
  Mila, McGill University\\
  \texttt{stephen.lu@mail.mcgill.ca} \\
  \And
  Ziqing Lu \\
  Genentech \\
  \texttt{lu.ziqing@gene.com} \\
  \And
  Ehsan Hajiramezanali \\
  Genentech \\
  \texttt{hajiramezanali.ehsan@gene.com} \\
  \And
  Tommaso Biancalani \\
  Genentech \\
  \texttt{biancalani.tommaso@gene.com} \\
  \And
  Yoshua Bengio \\
  Mila, Université de Montréal \\
  \texttt{yoshua.bengio@mila.quebec} \\
  \And
  Gabriele Scalia \\
  Genentech \\
  \texttt{scalia.gabriele@gene.com} \\
  \And
  Michał Koziarski  \\
  Mila, Université de Montréal \\
  \texttt{michal.koziarski@mila.quebec} \\
}

\begin{document}

\maketitle

\begin{abstract}
% Phenotypic screening allows us to discover novel therapeutics without necessarily understanding the underlying protein target. With the advent of machine learning, an approach to phenotypic discovery based on high-content imaging (HCI) became popular. Generative models for molecule generation can be guided by HCI data, typically with a computer vision algorithm trained in a supervised fashion to recognize phenotype of interest, and used as a reward function for generative model. However, this can require substantial amount of labeled data to train. In this paper we consider an alternative approach, in which we utilize unsupervised cross-modal contrastive learning. Specifically, we use GFlowNets trained with cross-modal latent similarity as a reward, and learn to generate new molecules that could produce similar phenotypic effect to the specified image target. We demonstrate that our method generates molecules with high morphological and structural similarity to the target, and increases the likelihood of them having a similar biological activity, as measured by an independent oracle model.

High-content phenotypic screening, including high-content imaging (HCI), has gained popularity in the last few years for its ability to characterize novel therapeutics without prior knowledge of the protein target. When combined with deep learning techniques to predict and represent molecular-phenotype interactions, these advancements hold the potential to significantly accelerate and enhance drug discovery applications. This work focuses on the novel task of HCI-guided molecular design. Generative models for molecule design could be guided by HCI data, for example with a supervised model that links molecules to phenotypes of interest as a reward function. However, limited labeled data, combined with the high-dimensional readouts, can make training these methods challenging and impractical. We consider an alternative approach in which we leverage an unsupervised multimodal joint embedding to define a latent similarity as a reward for GFlowNets. The proposed model learns to generate new molecules that could produce phenotypic effects similar to those of the given image target, without relying on pre-annotated phenotypic labels.  We demonstrate that the proposed method generates molecules with high morphological and structural similarity to the target, increasing the likelihood of similar biological activity, as confirmed by an independent oracle model.
\end{abstract}

\section{Introduction}

Historically, many drugs have been developed using the phenotypic screening approach, wherein the efficacy of a compound was assessed based on its observed biological effects without a prior comprehensive understanding of its underlying mode of action (MoA).
Although phenotypic discovery was gradually substituted or augmented by target-based discovery methods, it has experienced a resurgence in recent years, partly due to advancements in machine learning tools \cite{lin2020image,vincent2022phenotypic}.

High-content phenotypic screening, including high-content imaging (HCI) \cite{bray2016cell},  provides rich data such as morphological changes in cell shape and structure. This holds significant promise to aid discovery, serving as a rich and high-information readout of the perturbation's effect on the cell. Such readouts can potentially elucidate broad biological consequences, secondary effects, and MoAs \cite{moffat2017opportunities,vincent2022phenotypic}. Despite its potential to guide the drug discovery process, challenges persist in effectively utilizing this data. 

One common approach to utilize this information is training a supervised classification model to recognize the phenotype of interest (which, in turn, is extracted from the high-content readout), and then use it to virtually screen existing libraries \cite{krentzel2023deep}. However, this can require a substantial amount of labeled data and prior knowledge of the phenotype of interest, which might not be available when designing a therapeutic for a poorly understood disease. A preferred scenario would be to have a method capable of designing new compounds based on only a few, or even a single, target morphological readout. Another challenge lies in the screening process itself. Virtual screening limits the search space to existing screening libraries, which are significantly smaller than the whole drug-like space, often estimated to contain more than $10^{60}$ \cite{lipinski1997experimental} molecules.  This limitation can become particularly important when structurally and functionally novel molecules are desirable, for example, to improve the potency and diversity of the leads or to overcome unwanted secondary effects~\cite{jain2022biological,gao2024proteininvbench,song2023importance,ghari2023generative}.  % Here, generative models offer a promise of extending the search space size by orders of magnitude.

In this paper, we address both of these challenges by proposing a generative method guided by the target cell morphology.  Specifically, we propose a reward function that utilizes the latent similarity between the generated molecule and the target morphological profile. For this, we utilize a multimodal contrastive learning model that aligns small molecules to morphological profiles. Then, we use this reward to train a Generative Flow Network (GFlowNet)~\cite{bengio2023gflownet}, to generate molecules capable of inducing similar morphological outcomes. We demonstrate that our approach is capable of generating diverse molecules with high latent similarity to the provided morphological profile, which translates into a higher likelihood of obtaining similar predicted biological activity. We also show that the generation process can be structurally conditioned by using joint latent embeddings, which combine both target readout and molecule, further improving the performance. Practical use cases of the proposed method include generating molecules that induce a desired cell morphology obtained through gene perturbations \cite{rohban2022virtual},  scaffold hopping, where novel molecular structures with similar effects to a target one are desired \cite{hu2017recent}, and, more generally, molecular design guided by phenotypic readouts  \cite{krentzel2023deep}.

\section{Related work}

\textbf{Generative models for drug discovery.} There is a plethora of methods for molecular generation in the literature \cite{meyers2021novo,bilodeau2022generative}. They can be broadly categorized based on the molecular representation used: textual representation such as SMILES \cite{kang2018conditional,arus2020smiles,kotsias2020direct}, molecular graphs \cite{jin2018junction,maziarka2020mol,pedawi2022efficient,diamant2023improving} or 3D atom coordinate representations \cite{o20243d}; as well as the underlying generative methodology, e.g., variational autoencoders \cite{jin2018junction,maziarka2020mol}, reinforcement learning (RL) \cite{pedawi2022efficient,loeffler2024reinvent} or diffusion models \cite{runcie2023silvr}. These methods have found applications in drug discovery, reporting successes in several areas such as immunology and infectious diseases \cite{godinez2022design,moret2023leveraging}.
Recently, Generative Flow Networks (GFlowNets) \cite{bengio2021flow,nica2022evaluating,roy2023goal,shen2023tacogfn,volokhova2024towards,koziarski2024towards,gainski2024retrogfn,koziarski2024rgfn} emerged as a promising paradigm for molecular generation due to the ability to sample diverse candidate molecules, which is crucial in the drug discovery process. Importantly, similar to RL, GFlowNets can be trained based on the specified reward function, which makes them suitable for phenotypic discovery. Compared to existing methods, which focus on conditional generation based on a single property or multiple properties of interest, we tackle generation guided by high-content readouts, which we achieve through a multimodal latent joint representation.

\textbf{Deep learning for high-content molecular perturbations.}
High-content phenotypic screening, particularly high-content imaging (HCI), has become crucial in drug discovery for characterizing molecular effects in cells and elucidating targets, gene programs, and biological functions \cite{moffat2017opportunities,chandrasekaran2021image}. Recent advances propose deep learning techniques to accelerate and enhance these processes \cite{gavriilidis2024mini}.
Predictive models to infer the outcome of molecular effects have been developed both for transcriptomic readouts \cite{lotfollahi2019scgen,hetzel2022predicting,piran2024disentanglement} and HCI readouts \cite{palma2023predicting}. In these models, the output is highly multi-dimensional, capturing the full readout of the high-throughput experiment, thus potentially requiring a large amount of data for training and making it challenging to separate biological effects from background signals. In contrast to these works, we focus on the inverse problem, designing molecules leading to a specific (target) readout. 

\textbf{Joint representation of molecules and high-content readouts.}
Recent works focused on the joint modeling of molecules and high-content readouts. In particular, multimodal contrastive models \citep{radford2021learning} have been used to align molecular representations to high-dimensional readouts in latent space, thus capturing shared features while avoiding high-dimensional supervised losses \cite{sanchez2022contrastive,zheng2022cross,nguyen2023molecule,wang2023removing}. 
While these models can be used for screening tasks, reporting improved generalization ability compared to molecule-only models, they are unable to perform generative tasks. 
Compared to existing works in this area, we focus on the novel task of HCI-guided molecular design, while relying on a joint representation to guide the generation.
% sThis task requires navigating the extremely large and discrete chemical space through a high-dimensional readout, while having access to limited

\section{Method}

% TODO: graphical abstract

\subsection{Generative Flow Networks}

GFlowNets are amortized variational inference algorithms that are trained to sample from an unnormalized target distributions over compositional objects. GFlowNets aim to sample objects from 
 a set of terminal states $\mathcal{X}$  proportionally to a reward function $\mathcal{R} : X \rightarrow  \mathbb{R}^+$. GFlowNets are defined on a pointed directed acyclic graph (\textit{DAG}), $ G = (S, A)$, where:
 \begin{itemize}[leftmargin=.3in]
     \item $s \in S$ are the nodes, referred to as states in our setting, with the special starting state $s_0$ being the only state
with no incoming edges, and the terminal states $\mathcal{X}$ have no outgoing edges,
    \item $a = s \rightarrow s' \in A$ are  the edges, referred to as actions in our setting, and correspond to applying an action while in a state $s$ and landing in state $s'$. 
 \end{itemize}

We can define a non-negative flow function on the edges $F(s\rightarrow s')$ and on the states $F(s)$ of the DAG such that $\forall x \in \mathcal{X} F(x) = \mathcal{R}(x)$. A perfectly trained GFlowNet should satisfy the following flow-matching constraint:

\begin{equation}
    \forall s \in S\quad F(s) = \sum_{(s'' \rightarrow s) \in A} F(s'' \rightarrow s) = \sum_{(s \rightarrow s') \in A} F(s \rightarrow s' ).  
\end{equation}

A state sequence $\tau = (s_0 \rightarrow s_1 \rightarrow \dots \rightarrow
s_n = x)$, with $s_n=x \in \mathcal{X}$  and $a_i = (s_i \rightarrow s_{i+1}) \in A$  for all $i$, is called a complete trajectory. We denote the set of trajectories as $\Tau$.

\textbf{Trajectory balance.} There are several training losses that were explored to train a GFlowNet. Among those, trajectory balance \cite{malkin2022trajectory} has been shown to improve credit assignment. In addition to learning a forward policy $P_F$, we also learn a backwards policy $P_B$ and a scalar $Z_\theta$, such that, for every trajectory $\tau = (s_0 \rightarrow s_1 \rightarrow \dots \rightarrow
s_n = x)$, they satisfy: 

\begin{equation}
    Z_\theta \prod_{t=1}^{n} P_F(s_t|s_{t-1}) = R(x) \prod_{t=1}^{n} P_B(s_{t-1}|s_t) 
\end{equation}

\subsection{Multimodal contrastive learning}
\label{method:gmc}
Contrastive learning is a self-supervised approach that learns embeddings by maximizing agreement between similar samples and minimizing it between dissimilar ones, using contrastive loss functions like InfoNCE \cite{oord2018representation}. multimodal contrastive learning has emerged as a powerful approach for learning joint representations from diverse data modalities.  A notable instance of this approach is CLIP \cite{radford2021learning}, which aligns textual descriptions with visual representations, enabling robust cross-modal understanding. We leverage a multimodal contrastive model to learn a joint embedding of molecules and molecular effects. This choice avoids high-dimensional supervised losses and promotes learning ``informative'' features for the task (i.e., features that relate the two modalities to each other).

Let $\{(x_i, y_i)\}_{i=1}^N$ be a batch of $N$ pairs of molecular graphs $(x_i)$ and their corresponding morphology images $(y_i)$. Let $f$ and $h$ be the molecular and morphology encoders, respectively. Let the similarity between the molecular graph and morphology image embeddings be defined as \mbox{$S_{ij} = \exp \left( \frac{\text{sim}(f(x_i), h(y_j))}{\tau} \right)$}, where $\text{sim}(f(x_i), h(y_j))$ denotes the cosine similarity between embeddings.
The CLIP loss is defined as follows:
\begin{equation} \label{eq:clip}
L_{\text{CLIP}} = \frac{1}{N} \sum_{i=1}^N \left[ -\log \frac{S_{ii}}{\sum_{j=1}^N S_{ij}} - \log \frac{S_{ii}}{\sum_{j=1}^N S_{ji}} \right],
\end{equation}
where $\tau$ is a temperature parameter.

Instead of relying on CLIP, in this work, we leverage the closely related Geometric Multimodal Contrastive (GMC) loss \cite{poklukar2022geometric}. GMC leverages the same loss function in \eqref{eq:clip} to align modality-specific encoders to a joint encoder, which takes as input all modalities. Therefore, in addition to providing modality-specific embeddings $f$ and $h$, it also provides a \textit{joint} embedding $fh$ that we leverage when both the target molecule and its readout are available.

\subsection{GFlowNets for morphology-guided molecular design}
\label{ssec:method-descr}
% Question: should short primers be separate subsections (do we have enough content for this one then)???

\begin{figure}[!t]
    \centering
    \includegraphics[width=0.82\textwidth]{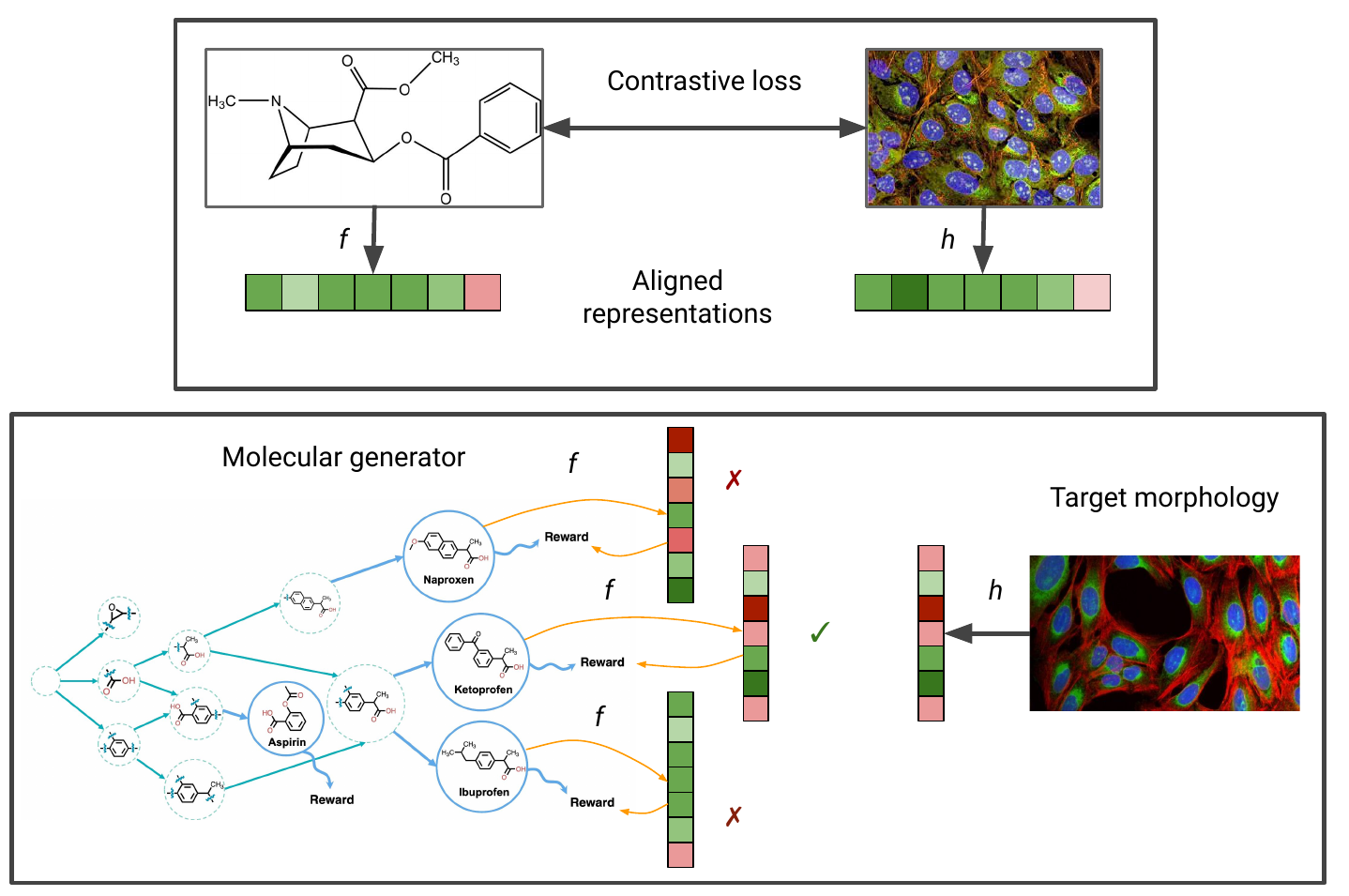}
    \caption{Overview of the proposed approach. In the first stage (top), cross-modal contrastive learning is used to train latent encoders $f$ and $h$ that produce aligned representations between molecules and readouts. Then, in the second stage (bottom), the target morphology readout is first converted into a latent vector, and the similarity between that and latents from molecules output by a generator is used as a reward. The model learns to sample molecules capable of inducing similar latents to the target.}
    \label{fig:abstract}
\end{figure}

The proposed approach combines recent developments in multimodal contrastive learning and molecular generation with GFlowNets into a unified pipeline, as illustrated in \Cref{fig:abstract}. The method relies on first training a contrastive learning model capable of producing aligned latent representations between chemical structures and morphology profiles, and then using these representations as a guiding signal for GFlowNet. We define the reward function for the GFlowNet by training a GMC model as described in Section~\ref{method:gmc}. Specifically, we use the following reward:
\begin{equation}
    R(x|y) = 1 + \frac{f(x)h(y)}{2\lVert f(x)\rVert \lVert h(y)\rVert},
\end{equation}
which is a normalized cosine similarity between the target morphology latent and the generated molecular latent. This choice is important to enforce the non-negativity of the GFlowNet reward. During the training of the GMC model, we impose early stopping using the correlation between the cross-modality distance metric. We observe that early stopping reduces the variance of cosine similarity between the multimodal GMC embeddings, while it does not affect the performance of GFlowNet (see \Cref{apx:early-stopping} for more details). Inspired by using replay buffer in reinforcement learning~\cite{vemgal2023an}, we leverage replay buffer with known decomposed structure when training on joint morphology and structure-guided generations. This increases the structural similarity of generated samples to the known target.

\section{Experimental study}

% We first verify the underlying assumption of our method, namely that the similarity in latent space produced by the contrastive model correlates with the morphological similarity. Afterwards, we evaluate the proposed approach by examining the quality and diversity of generated samples.  Next, we consider the setting of optimization biased towards known perturbation, where we base our target latent on the combination of target morphology and the associated molecular structure. Finally, we examine predicted biological activity of generated samples in downstream tasks.

We first verify the underlying assumption of our method, namely that the latent similarity derived through contrastive learning correlates with the morphological similarity. Afterwards, we evaluate the proposed approach, first by examining the quality and diversity of the generated samples, and then by analyzing their predicted biological activity in downstream tasks. Finally, we consider the setting of structurally conditioned generation, where we define the target latent based on the combination of the target morphological profile and associated molecular structure, with the goal of biasing the generation toward a known perturbagen.

\subsection{Set-up}
\label{exp:setup}

% TODO:
We perform experiments on the Cell Painting dataset introduced by \cite{bray2016cell,bray2017dataset}, as further pre-processed in \cite{moshkov2023predicting}. The dataset includes 16,170 molecules and associated cell morphology images. Each image includes five color channels that describe the morphology of five cellular compartments: nucleus (DNA), Endoplasmic reticulum (ER), nucleolus/cytoplasmic RNA (RNA), F-actin cytoskeleton (AGP), and Mitochondria (Mito). Images have been pre-processed using CellProfiler 3.0 \cite{mcquin2018cellprofiler,moshkov2023predicting}. 
% dataset description

Additionally, to support the evaluation of the generated molecules, we leverage oracle models independently trained on data from multiple assays released by the Broad Institute \cite{moshkov2023predicting}. For this, we primarily focus on assays that have been linked to morphological features and/or combinations of morphological and chemical properties. We select 37 assays identified in  \cite{moshkov2023predicting} as predictable from morphological features or combined chemical and morphological features with high accuracy (AUROC $> 0.9$). This allows us to evaluate the ability of the model to generate molecules with biochemical and cellular effects similar to the (unknown) target molecule.

% TODO
% hyperparameters in appendix
% how targets were selected

\subsection{Relation between proposed reward and morphological distance}
% Working title
% correlation plots

The underlying assumption behind our method is that the latent representations produced by the contrastive learning model are of a high enough quality to reliably compute similarity. Specifically, given a target pair of a molecule and its associated morphology $(\dot{x}, \dot{y})$, an arbitrary pair of a generated molecule and the morphology that would be induced by that molecule $(x, y)$, distance $d$ in the input modality space, distance $\hat{d}$ in the latent space, and models $f$ and $h$ that transform the inputs into a latent representation, we require that 
\begin{equation}
    R(x|\dot{y}) = \hat{d}(h(\dot{y}), f(x)) \sim d(\dot{y}, y).
\end{equation}
It is worth noting that during generation, we always know $x$, which is the molecule generated by our model, and never know $y$, which is why we introduce the encoder $f$. Similarly, we always know the target morphology $\dot{y}$, and in some settings, such as finding drug analogs, we might also have access to the associated molecule $\dot{x}$. In particular, in the latter setting, we might consider conditioning on the joint latent, produced based on the pair of $(\dot{x}, \dot{y})$: $R(x|\dot{x}, \dot{y}) = \hat{d}(fh(\dot{x}, \dot{y}), f(x))$ (see Section~\ref{ssec:joint}).

% \begin{equation}
%     R(x|\dot{x}, \dot{y}) = \hat{d}(f(\dot{x}, \dot{y}), f(x)) \sim d(\dot{y}, y).
% \end{equation}

\begin{figure}[!htb]
    \centering
    \includegraphics[width=.9\textwidth]{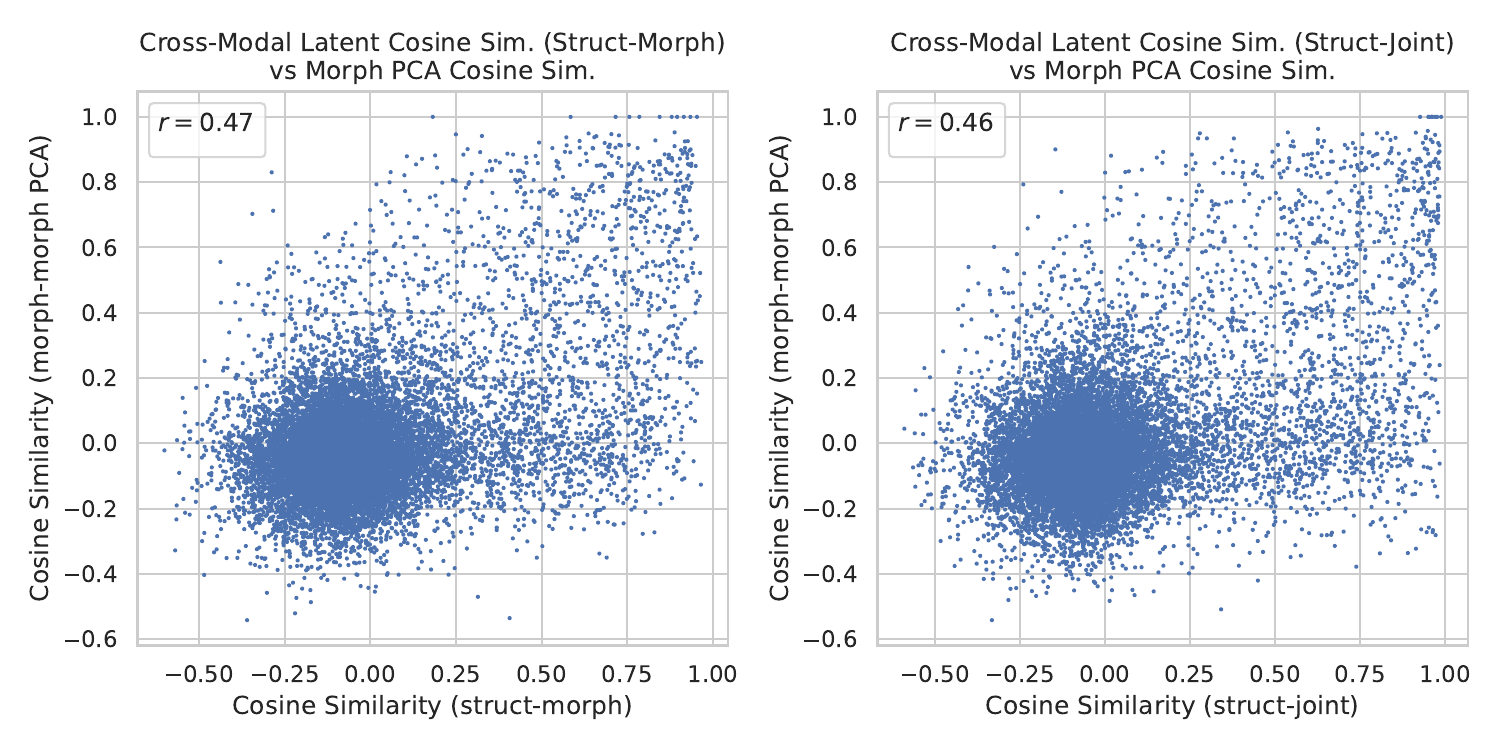}
    
    \caption{Correlation between the similarity in morphological feature space and latent similarity between target structure and a) morphology (left) or b) joint morphology and structure (right).}
    \label{fig:corr}
\end{figure}

To evaluate this assumption, for every pair of observations from the dataset, we measure the correlation between the similarity of morphological features and the similarity between the latent representations of the structure from the first observation and the target morphology from the second observation. Additionally, we also measure the same for the joint target latent, computed based on both the morphology and the associated molecule structure. The results are presented in \Cref{fig:corr}. As can be seen, in both cases we observe a medium level of correlation. While not perfect, likely due to inherent variance in the morphological profiles and model uncertainty, we argue that this can be sufficient for screening purposes. % (where we are interested in improving the hit ratio, but due to the difficulty of the task do not expect very precise outcomes).

\subsection{Generating high reward and diverse samples}
% Working title

We are interested in evaluating the capabilities of a GFlowNet in optimizing the specified reward function. Given our primary focus on its application in the initial discovery stage, it is essential to generate not only high-reward outcomes but also a diverse set of samples. We benchmarked the proposed approach against random sampling (RND) and two standard RL-based baselines: soft Q-learning (SQL) \cite{haarnoja2017reinforcement} and soft actor-critic (SAC) \cite{haarnoja2018soft}. Note that since, to the best of our knowledge, this is the first published attempt at guiding the generative molecular model with expected image morphology outcome, we focus specifically on benchmarking against other potential molecular generation methods.

\begin{figure}[!htb]
    \centering
    % distribution plots of rewards
    \includegraphics[width=0.49\textwidth]{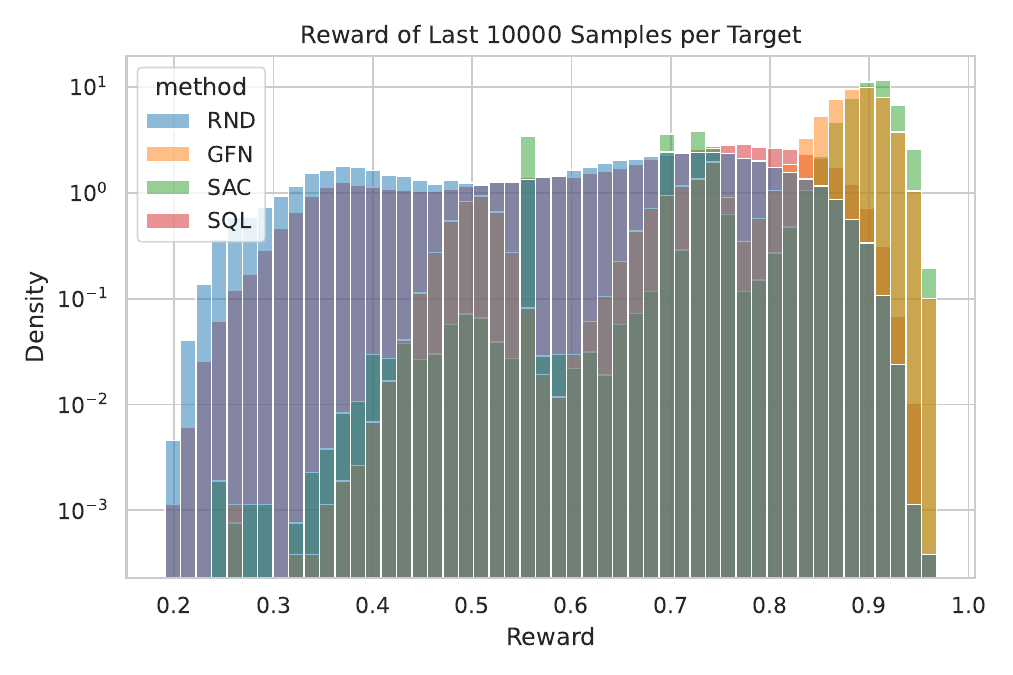}
    % number of modes found
    \includegraphics[width=0.49\textwidth]{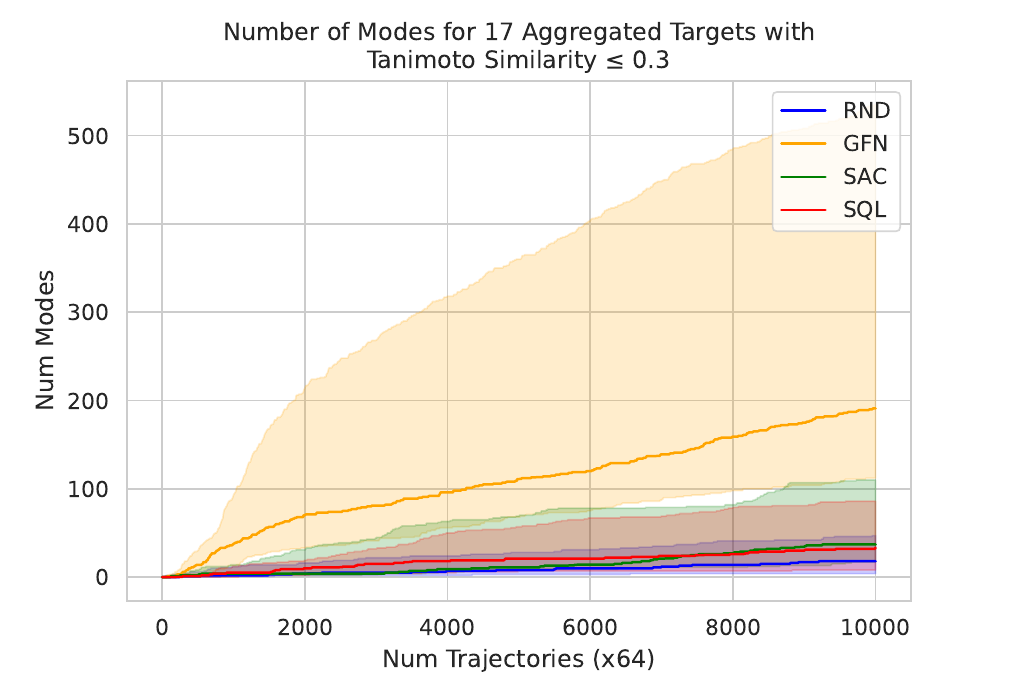}
    \caption{Comparison of examined methods in terms of reward optimization: a) distribution of rewards from generated samples, 10,000 samples generated by a trained model (left) and b) number of modes discovered (right). Both plots aggregate across all examined targets.}
    \label{fig:morph-reward-modes}
\end{figure}

\begin{figure}[!htb]
    \centering
    % Tanimoto similarity distribution (between modes)
    \includegraphics[width=0.49\textwidth]{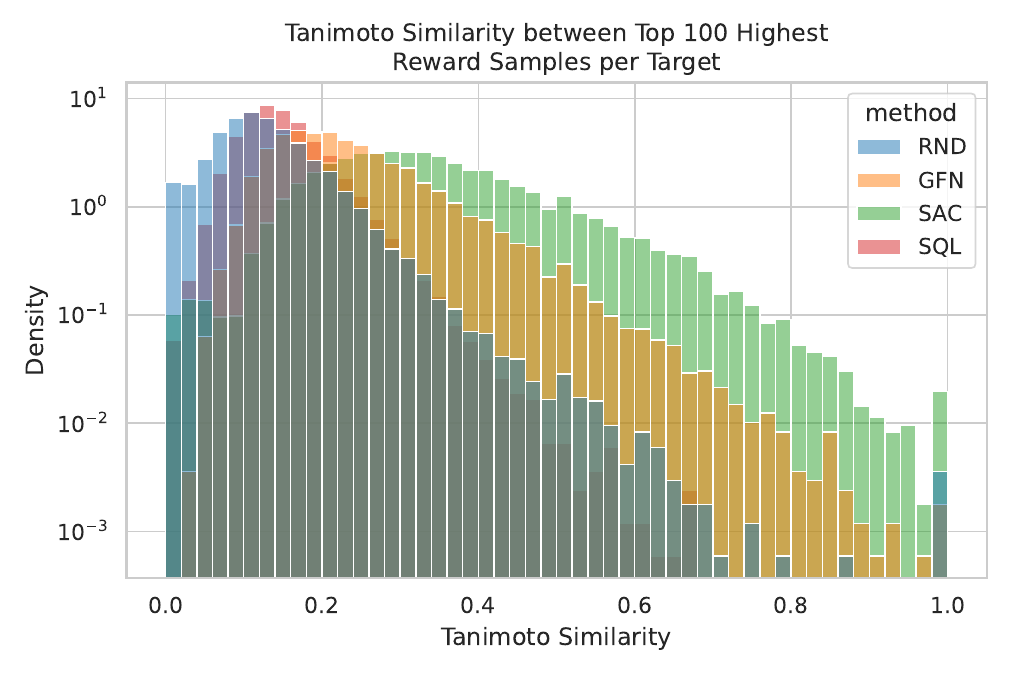}
    % UMAP vs. reference methods (that shows diversity)
    \includegraphics[width=0.49\textwidth]{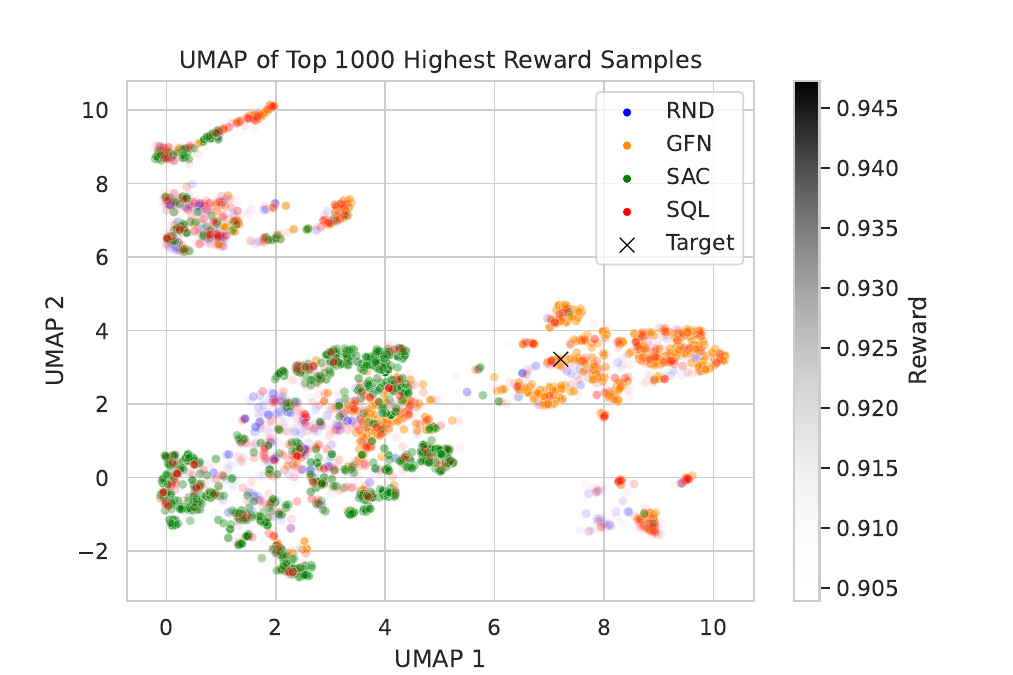}
    \caption{Comparison of examined methods in terms of diversity: a) distribution of Tanimoto similarities between top-100 generated samples (left; aggregated across all examined targets, lower is better) and b) structural diversity of top-1000 highest reward samples to Target \#8636 (right).}
    \label{fig:morph-diversity}
\end{figure}

We show the distribution of rewards for generated samples and the number of discovered modes (defined as molecules with reward $\ge$ 90th percentile and Tanimoto similarity to other modes $\le 0.3$) in \Cref{fig:morph-reward-modes}, and the distribution of similarities between top-100 generated samples in \Cref{fig:morph-diversity} (each figure aggregated across all considered targets). As shown, GFlowNet learns to sample high-reward candidate molecules (with a significantly higher average reward than random sampling and SQL, comparable to SAC) while also significantly improving the diversity compared to SAC (with a lower similarity between top candidates). Both of the above translate into a significantly higher number of discovered modes than the baseline methods. The diversity of the generated samples is further illustrated in \Cref{fig:morph-diversity}, where a UMAP visualization of the molecular structures produced by different methods is presented for a specific target. As can be seen, GFlowNet displays sample coverage similar to random sampling, which is a desirable outcome.

% TODO:

\subsection{Biological activity estimation}
% Working title, oracle-based evaluation

So far, we have established that the proposed approach is capable of generating diverse samples with high reward and that there is a moderate correlation between the reward and the similarity to the target. However, what is critical in the end is whether this will translate into generated molecules inducing similar biological effects to the original target perturbation (ground truth). Ideally, we wish to evaluate this on the basis of experiments comparing generated molecules and the ground truth, but this can be costly and time-consuming. Instead, in the following,  we estimate the similarity of the biological effect based on the available data. We consider two approaches to this goal.

\begin{figure}[!htb]
    \centering
    \includegraphics[width=0.49\textwidth]{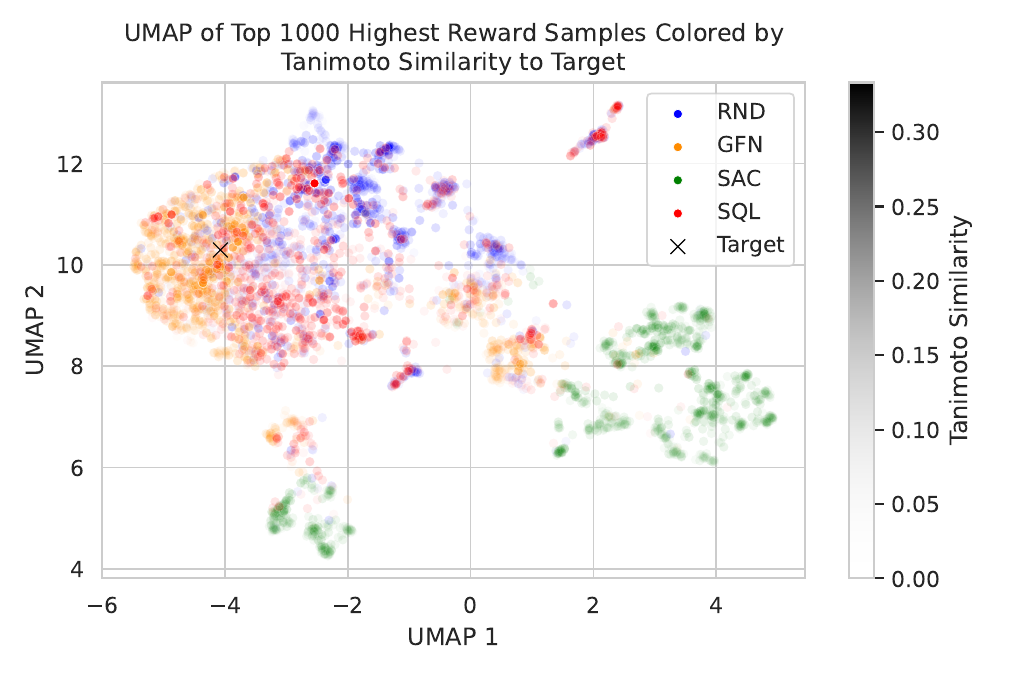}
    \includegraphics[width=0.49\textwidth]{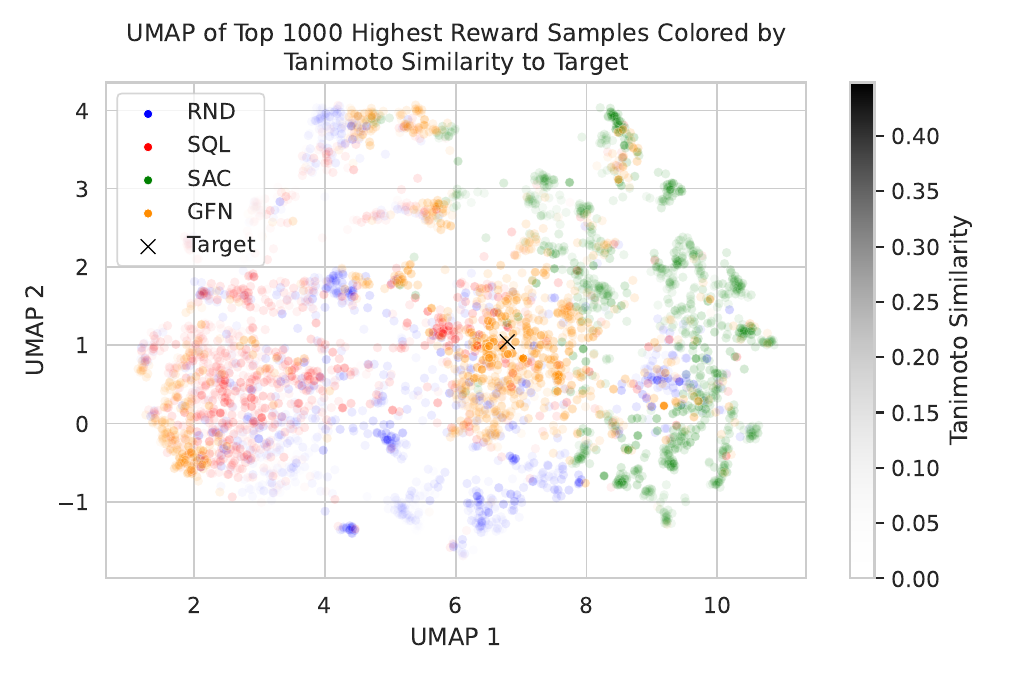}
    \caption{UMAP of top-1000 highest reward samples for a) morphology-only target (left) and b) joint target (right). Colored by Tanimoto similarity to target \#2288.}
    \label{fig:umap_sim_to_target}
\end{figure}

First of all, we would expect a perfectly optimized generator to be able to sample known molecules that induced target morphological profiles. In practice, this might be unattainable: not only is the problem itself heavily under-constrained (we expect multiple molecules, likely a very large number, to be able to induce a given morphological outcome), but also our morphological similarity estimation is intrinsically noisy. Because of the above, what we try to achieve is the highest possible structural similarity of generated samples to the known molecule that induced the given morphology. The maximum Tanimoto similarity to the known molecular target, averaged across all considered targets, is presented in \Cref{tab:max_sim}. 
As shown, the proposed approach generally recovers the underlying targets more effectively, suggesting its utility in identifying molecules with expected biological activity. Unsurprisingly, this effect is particularly pronounced when conditioning on specific molecular structures. Target recovery is further illustrated in \Cref{fig:umap_sim_to_target}, where we demonstrate structural proximity of the generated samples to the known target (visualizations for other targets can be found in \Cref{apx:umap}).

% TODO:
% UMAP with similarity to the target

% table with average maximum Tanimoto similarity to the actual target
\begin{table}[!t]
    \caption{Max. Tanimoto similarity to the target in last 10,000 samples, averaged across all targets.}
    \label{tab:max_sim}
    \centering
    \begin{tabular}{ccccc}
    \toprule
       & Random & Soft Actor-Critic & Soft Q-Learning & GFlowNet \\
       \midrule
       Morphology Target & 0.305 (± 0.057) & 0.261 (± 0.068) & 0.329 (± 0.079) & \textbf{0.337 (± 0.092)} \\
       Joint Target & 0.311 (± 0.064) & 0.388 (± 0.163) & 0.309 (± 0.064) & \textbf{0.451 (± 0.163)} \\
    \bottomrule
    \end{tabular}
\end{table}

\begin{figure}[!htb]
    \centering
    \includegraphics[width=0.49\textwidth]{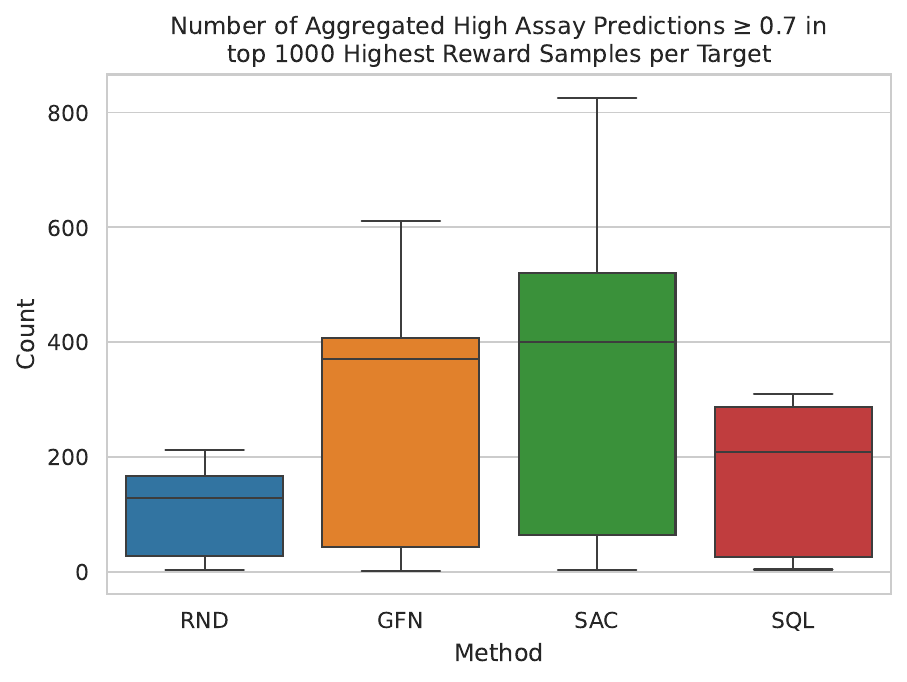}
    \includegraphics[width=0.49\textwidth]{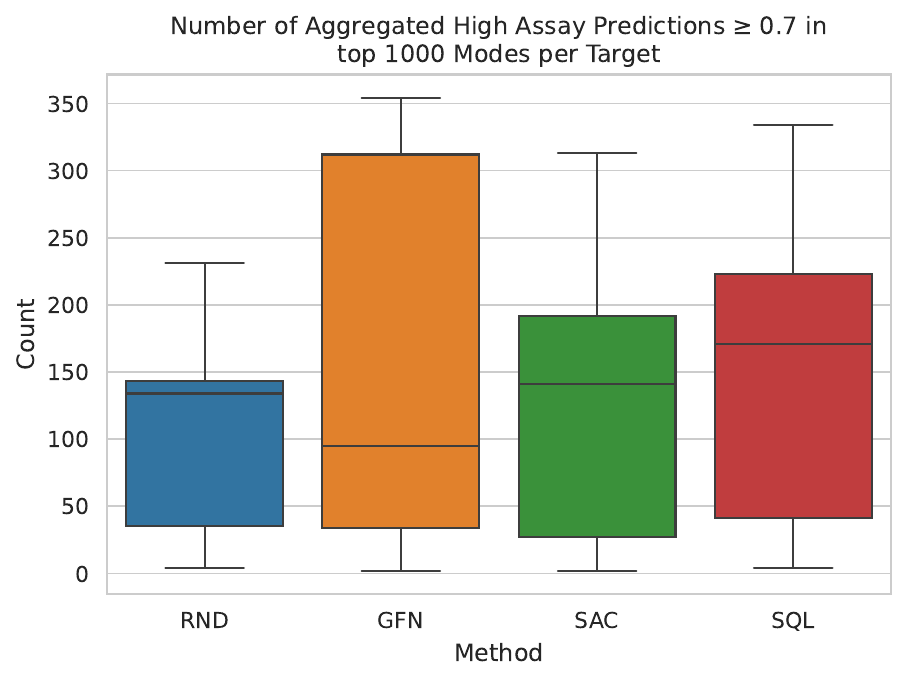}
    \caption{Number of samples with predicted assay probability $\ge$ 0.7 in a) top 1000 highest reward samples (left) and b) top 1000 modes with Tanimoto similarity $\le$ 0.3 (right). Results are aggregated across multiple targets.}
    \label{fig:assay_preds}
\end{figure}

The second approach we consider involves utilizing oracle models for predicting biological activity. We train an MLP using molecular structures, specifically their extracted molecular fingerprints, as inputs to predict the outcomes of biological assays (details of the model training are provided in \Cref{apx:oracle}). For each target molecule, at least one associated assay has a positive outcome. The objective is to generate molecules with a high predicted probability of having a positive outcome in the specific assay for which the target molecule has known activity.
The number of generated samples with high predicted assay probability ($\ge 0.7$) is presented in \Cref{fig:assay_preds}.
We present the results for both the top-1000 highest reward samples per target, as well as 1,000 modes per target.
Note that, due to the high uncertainty of the oracle model, we are primarily interested in quantifying the number of high-likelihood samples. % to increase the reliability of the prediction.
As can be seen, using guided generation helps improve the proportion of generated molecules with high predicted assay probability, which serves as a proxy for generating molecules with similar biological activity. It is worth noting that while we observe a higher predicted probability for SAC when considering the top-1000 molecules by reward, this trend reverses when considering different modes, once again highlighting the higher diversity combined with the high reward of GFlowNet samples. It is also worth noting that we observe a large variance across targets. While this can be partially attributed to the uncertainty of the oracle, further investigation of factors determining whether the proposed approach is helpful or not for a given target is an important future research direction.

\subsection{Joint morphology and structure-guided generation} \label{ssec:joint}

Until now, we focused on the generation of molecules capable of inducing a specific morphological outcome, assuming that the perturbation that led to the observed outcome is fully unknown. 
However, it is worth noting that in practice, this problem can be significantly under-constrained, as there might be a large number of molecules resulting in similar morphological effects, and many of them could have undesirable properties (e.g. toxicity, lack of stability, etc.).
There are multiple ways of constraining generated molecules to a particular chemical subspace with desirable properties, for instance, by including additional reward terms and conditions in the GFlowNet training. However, a particularly well-suited approach for our method, assuming the availability of the molecular structure associated with the target morphology (e.g. in the drug analog search setting), involves conditioning on the joint latent representation. This strategy aims to anchor the generated molecules to a known, desirable molecular structure.

\begin{wrapfigure}{R}{.45\columnwidth}
    \centering
    \vspace{-.5cm}\includegraphics[width=0.45\textwidth]{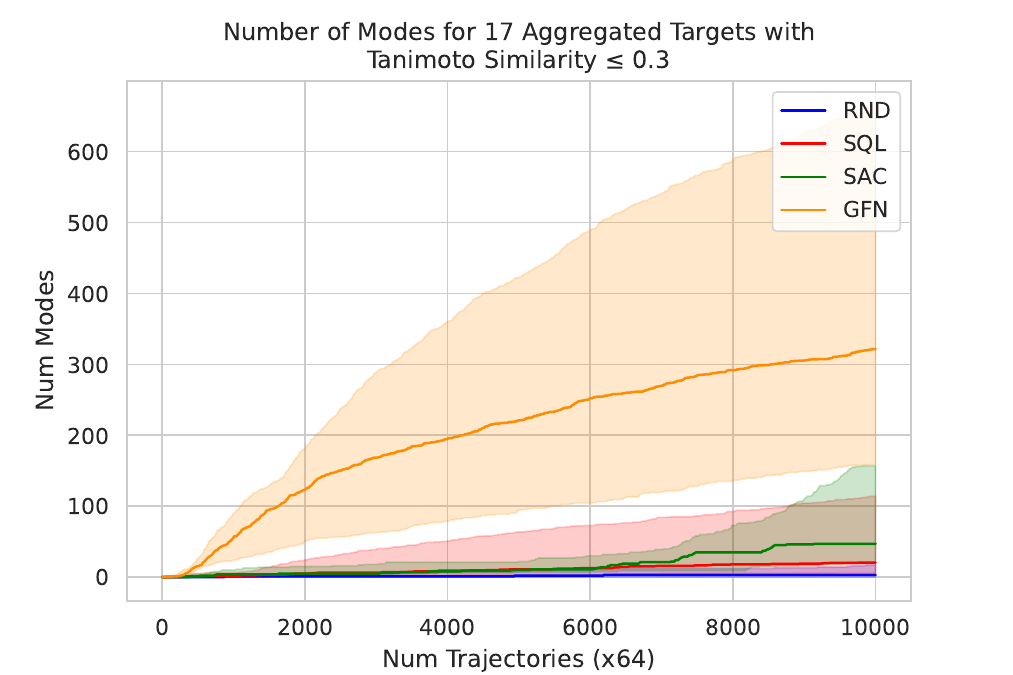}
    \caption{Number of modes discovered for joint morphology and structure-guided generation.}
    \label{fig:joint_num_modes_over_trajs}
\end{wrapfigure}

\begin{figure}[!t]
    \centering
    \includegraphics[width=0.42\textwidth]{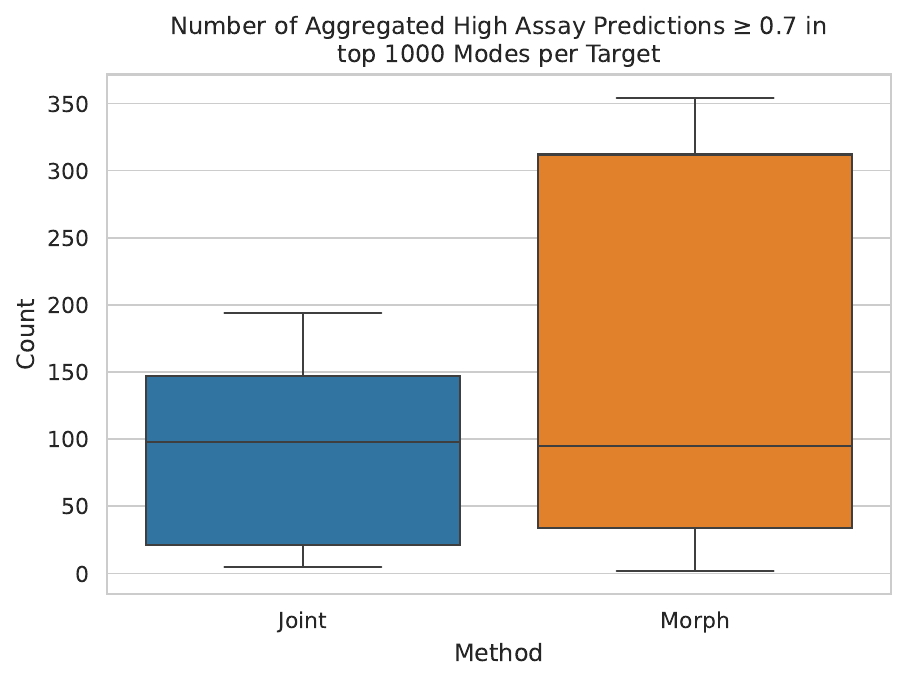}
    \includegraphics[width=0.42\textwidth]{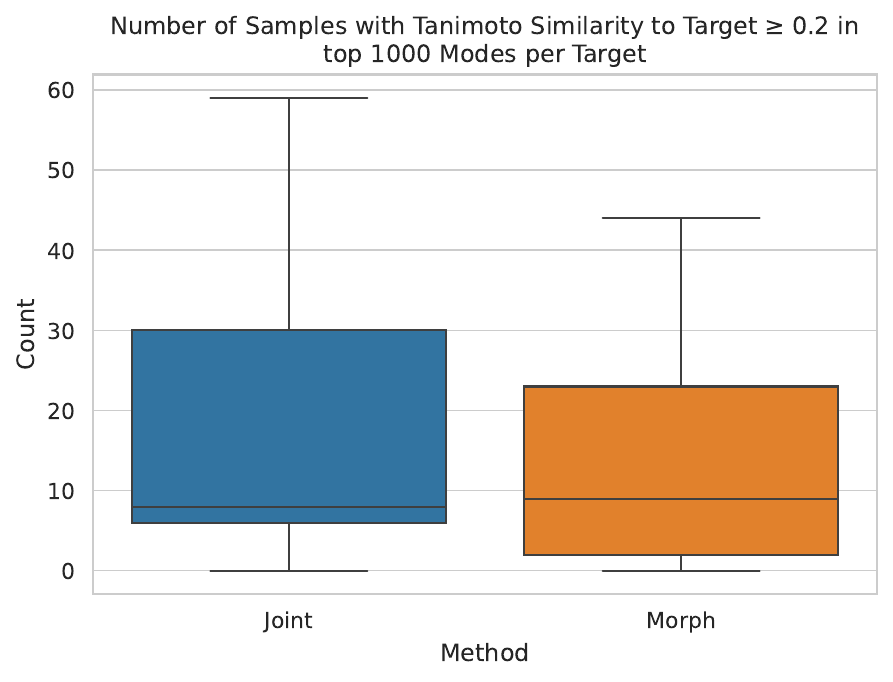}
    \caption{Comparison of morphology-only versus joint morphology and structure guided generation: a) number of samples with prediction assay probability $\ge$ 0.7 in top 1000 modes (left; higher is better) and b) number of samples in top 1000 modes with Tanimoto similarity to target $\ge$ 0.2 (right). Modes are defined as high reward samples with mutual Tanimoto similarity $\le$ 0.3.}
    \label{fig:joint_vs_morph}
\end{figure}

We evaluate the capabilities of the proposed approach in constraining the searchable space based on the given structure by replacing the target latent with a joint representation generated based on both morphology and associated structure (Section~\ref{method:gmc}). We evaluate the number of discovered modes (\Cref{fig:joint_num_modes_over_trajs}), molecular similarity to the given target (\Cref{tab:max_sim}), and number of samples with high predicted assay probability (\Cref{fig:joint_vs_morph}). As can be seen, using the joint representations does not impact the ability of GFlowNet to generate high-reward and diverse samples, and increases the average number of discovered modes. As expected, it also leads to generating compounds more structurally similar to the target molecules, effectively constraining the generation process. However, somehow surprisingly, conditioning based on the joint representation does not seem to increase the proportion of molecules with high assay predicted probability.

Our leading hypothesis is that the joint embedding space learned by the GMC model encodes a stronger structural signal than a morphological signal, thus leading the GFlowNet to sample molecules that are more similar to the target, but that don't necessarily trigger the desired morphological profile in the target cell. This claim is supported by \Cref{apx:gmc_single}, where the GMC alignment between joint and structural latent space ($r=0.94$) is greater than the alignment between joint and morphology ($r=0.88$). Although the alignment between structure and morphology embeddings is even lower ($r=0.75$), we argue that this setting provides a stronger and more direct signal that guides the GFlowNet towards more diverse molecules that exhibit the desired morphological profile of the target. This is consistent with our results in \Cref{fig:joint_vs_morph}, which suggest that the joint setting yields a higher number of molecules that are structurally similar to the target but finds fewer molecules with high assay probabilities than the morphology-only setting.\footnote{Code available @ \url{https://github.com/TheMatrixMaster/omics-guided-gfn}}

% TODO:

% Working title, joint embedding
% TODO: joint embedding results

\section{Limitations}
\label{sec:limit}

The current approach heavily relies on the quality of the learned latent representations. As it has been shown, the current multimodal contrastive models provide only a medium correlation between latent and morphology similarities. This makes the approach suitable for larger screening campaigns, but would likely lead to high variance with smaller screening budgets. However, the proposed framework is compatible with different multimodal learning frameworks, and we expect that current research on more robust joint representations \cite{wang2023removing} will lead to more efficient and precise generation capabilities.

On the generative model side, the proposed approach does not take into account non-primary molecular properties, such as synthesizability, toxicity, drug-likeness, etc. However, while not considered in this work, this can, in principle, be integrated by adding additional reward terms \cite{korablyov2024generative}. Additionally, in the current state, even if leveraging the same pretrained multimodal model, the approach requires training a separate model for each considered morphology target. Further work on developing a conditional variant, where a single GFlowNet samples conditionally given a morphology readout, would be required to reduce the computational cost of the method. % Additional experiments, including significantly higher number of morphology targets and wet lab screens, would increase the reliability of evaluation.

\section{Conclusion}

In this paper, we consider the task of molecular generation guided by a high-dimensional morphological profile. Such a framework is broadly applicable, for example, in designing drugs mimicking the effect of a genetic perturbation, designing drug analogs, or, more generally, molecular design guided by phenotypic readouts. The proposed approach relies on the GFlowNet framework for molecular generation, and uses a reward based on the latent similarity of representations from a multi-modal contrastive learning model. To the best of our knowledge, this is the first published attempt at the challenging task of guiding the generative molecular model with the expected image morphology outcome. We experimentally demonstrate the usefulness of the proposed approach for generating diverse drug candidates, which was shown to increase the likelihood of producing molecules with similar biological activity when compared to random screening.

Future directions include additional evaluation of the generated molecules, in particular lab experiments. This could also support an active learning process, with the aim of improving the joint representation through the acquisition of new data. Potential extensions to the model include making the approach conditional on the provided morphology target, and further constraining the optimization with respect to multiple properties of interest. 

% Potential positive societal impacts mostly include facilitating the process of drug discovery and finding novel medications. A potential negative impact includes, e.g., facilitating the development of purposefully toxic compounds.

\begin{ack}
The research was supported by funding from CQDM Fonds
d’Accélération des Collaborations en Santé (FACS) / Acuité Québec and Genentech. Z.L., E.H., T.B., and G.S. are employees of Genentech, Inc., and shareholders of F. Hoffmann-La Roche Ltd.
\end{ack}

\bibliography{references}

%%%%%%%%%%%%%%%%%%%%%%%%%%%%%%%%%%%%%%%%%%%%%%%%%%%%%%%%%%%%

\appendix

\section{Additional experiments and plots}
\label{apx:experiments}

\subsection{Target selection}
\label{apx:targets}
\begin{figure}[!htb]
    \centering
    \includegraphics[width=0.44\textwidth]{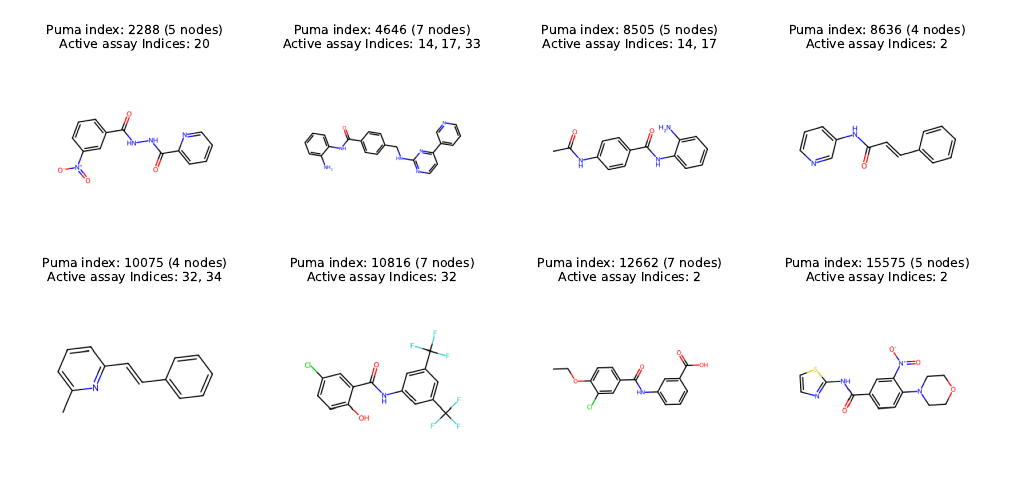}
    \includegraphics[width=0.55\textwidth]{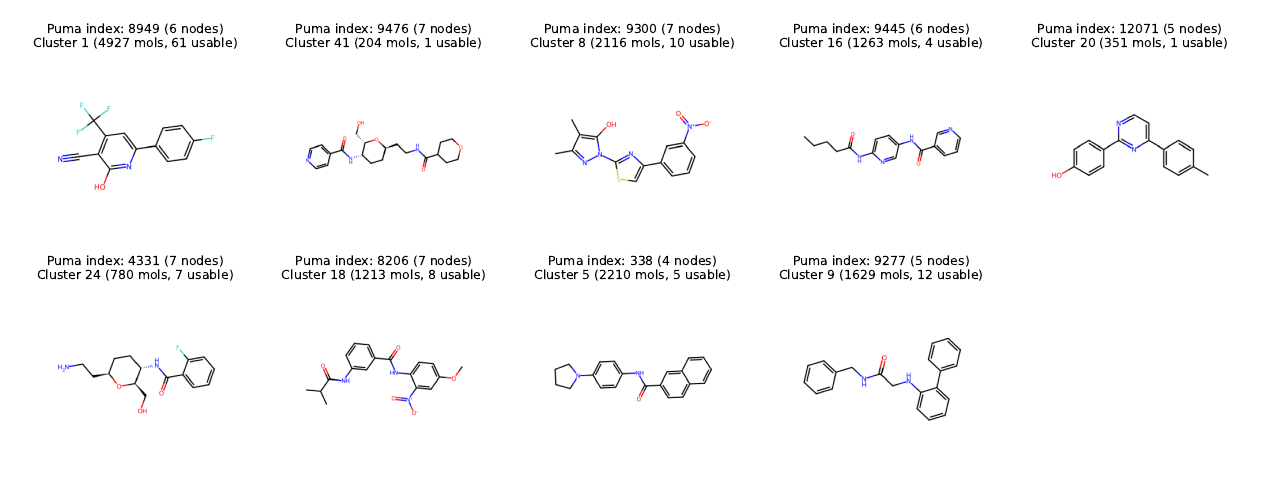}
    \caption{Set of targets used for our experiments, 8 of which are biologically active bassed on the ground truth assay information, and were used for oracle evaluation (left group).}
    \label{fig:targets}
\end{figure}

To evaluate the generative models, we selected a total of 17 targets from the Cell Painting dataset, among which 8 were used for the biological assay activity experiments (\Cref{fig:targets}). All targets were verified to be decomposable into the molecular fragment set used for generative modelling and were chosen to represent a diverse set of active biological assays for the oracle-based evaluation.

\subsection{GMC model selection}
\label{apx:early-stopping}
For GMC model selection, we tried early stopping on both the multimodal contrastive loss in \ref{eq:clip} and the cross-modality correlation metric presented in Section~\ref{ssec:method-descr}. While there was not a significant difference in GFlowNet performance between these two variants, we found that early stopping on the correlation metric reduced the variance on the cosine similarity between molecular embeddings and embeddings of their associated morphological or joint readouts. In other words, the cross-modality agreement learnt by this model had lesser spread.

\begin{figure}[!htb]
    \centering
    \includegraphics[width=\textwidth]{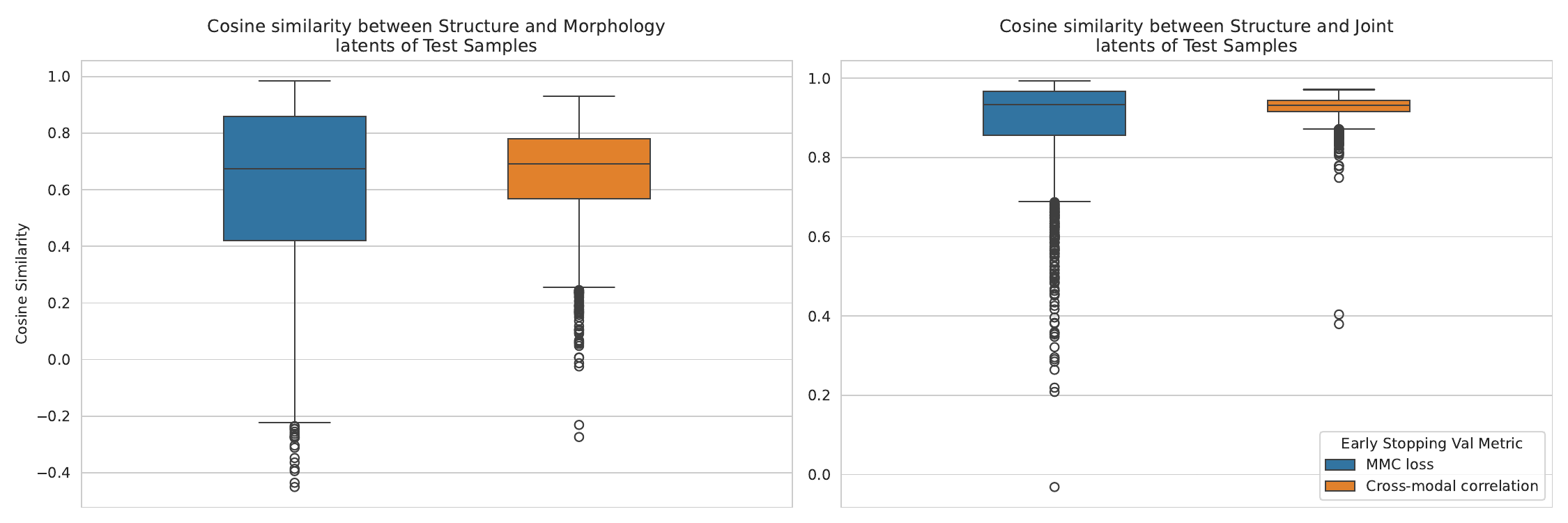}
    \caption{Cosine similarity between a) structure and morphology embeddings (left) and b) structure and joint embeddings (right) produced by GMC model on test split.}
    \label{fig:mmc_earlystop}
\end{figure}

\subsection{Additional UMAP plots per target}
\label{apx:umap}
We produce UMAP plots of the top 1000 highest reward molecules for morphology-only and joint morphology and structure-guided generation. We then color these samples by reward and Tanimoto similarity to the target, such that higher values (which are more desirable) have higher opacity. We use Morgan fingerprints with a radius of 3 and a dimensionality of 2048. UMAP representations are computed with 30 neighbors, minimum distance 0.1, and Jaccard metric. Below, we include the results for all targets in \Cref{apx:targets}. The first two columns consist of the samples obtained for morphology-only generation, while the last two columns consist of samples obtained for joint morphology and structure-guided generation.

% \vspace{1.5em}
\rotatebox{90}{\hspace{1.5em}\# 2288}
\includegraphics[width=0.24\textwidth]{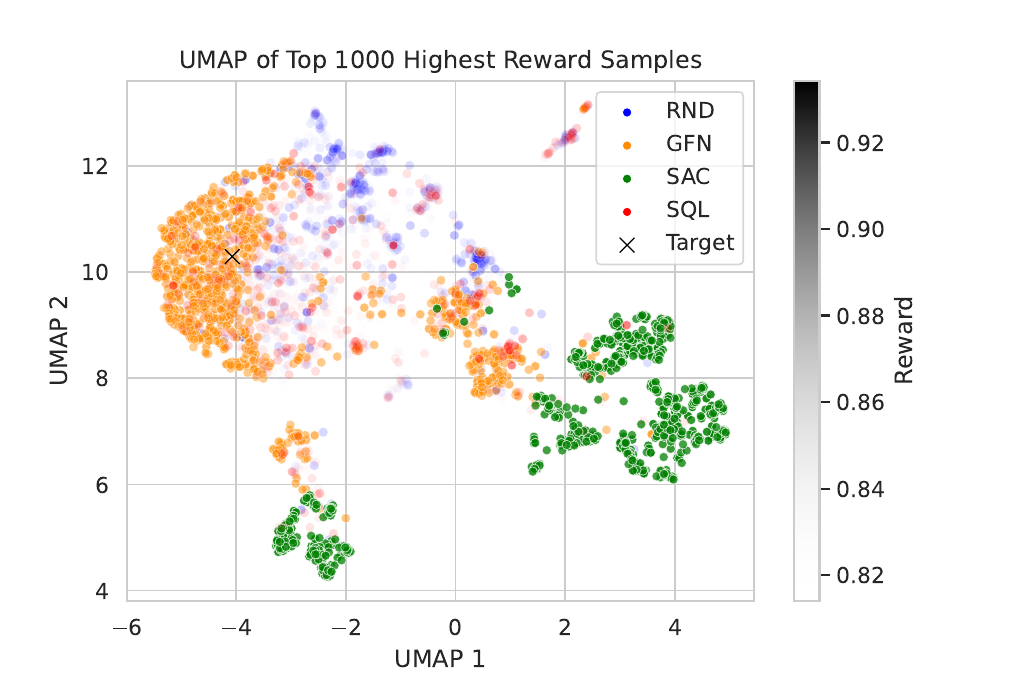}
\includegraphics[width=0.24\textwidth]{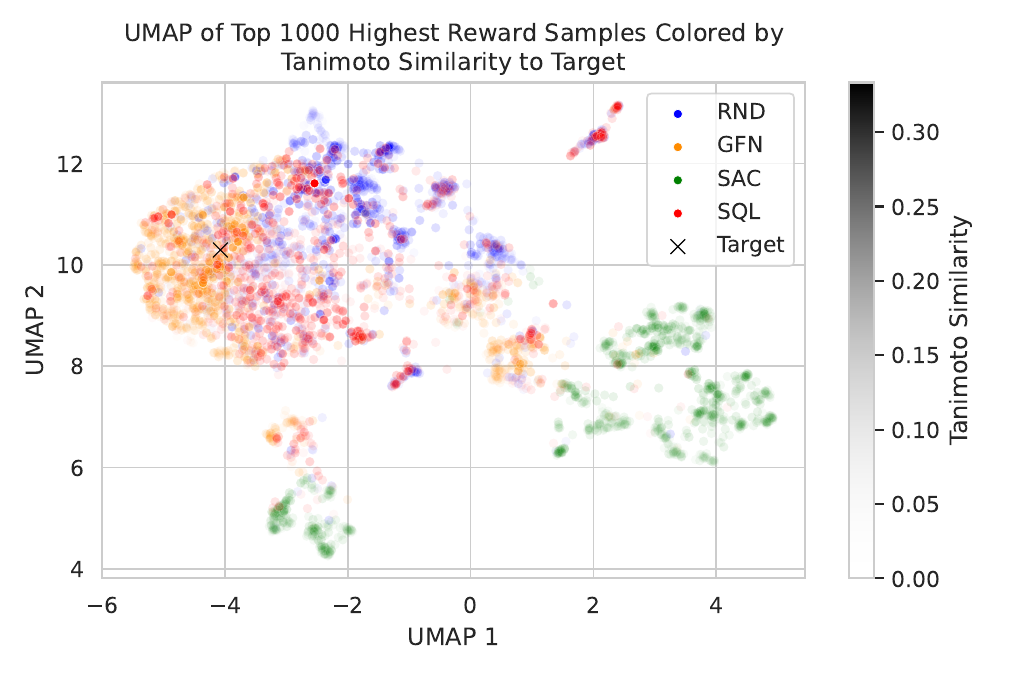}
\includegraphics[width=0.24\textwidth]{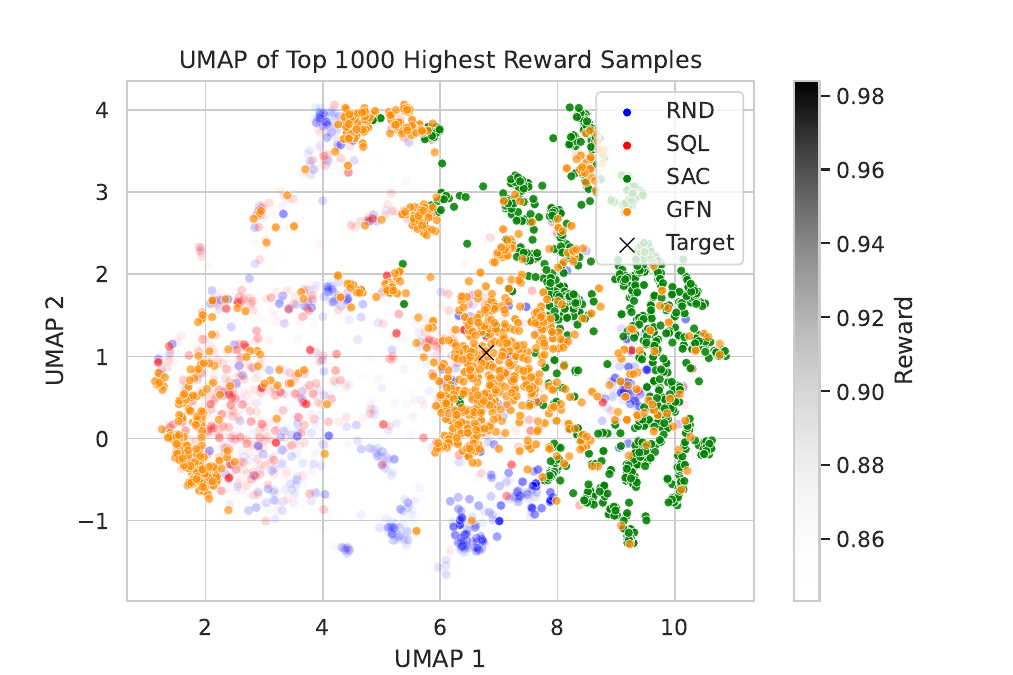}
\includegraphics[width=0.24\textwidth]{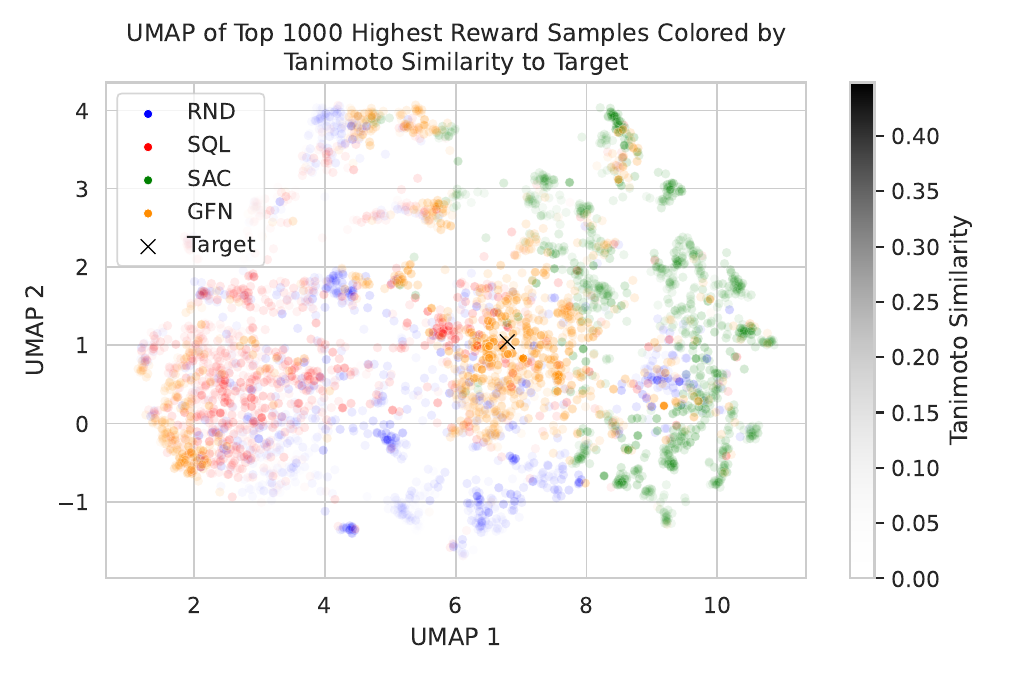}

\rotatebox{90}{\hspace{1.5em}\# 4646}
\includegraphics[width=0.24\textwidth]{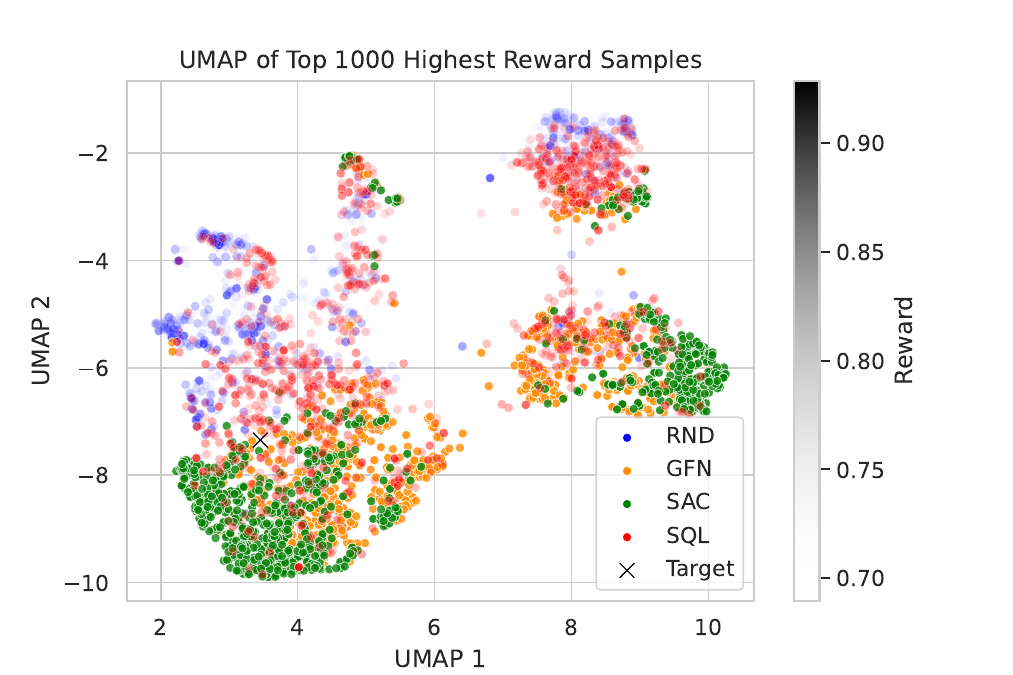}
\includegraphics[width=0.24\textwidth]{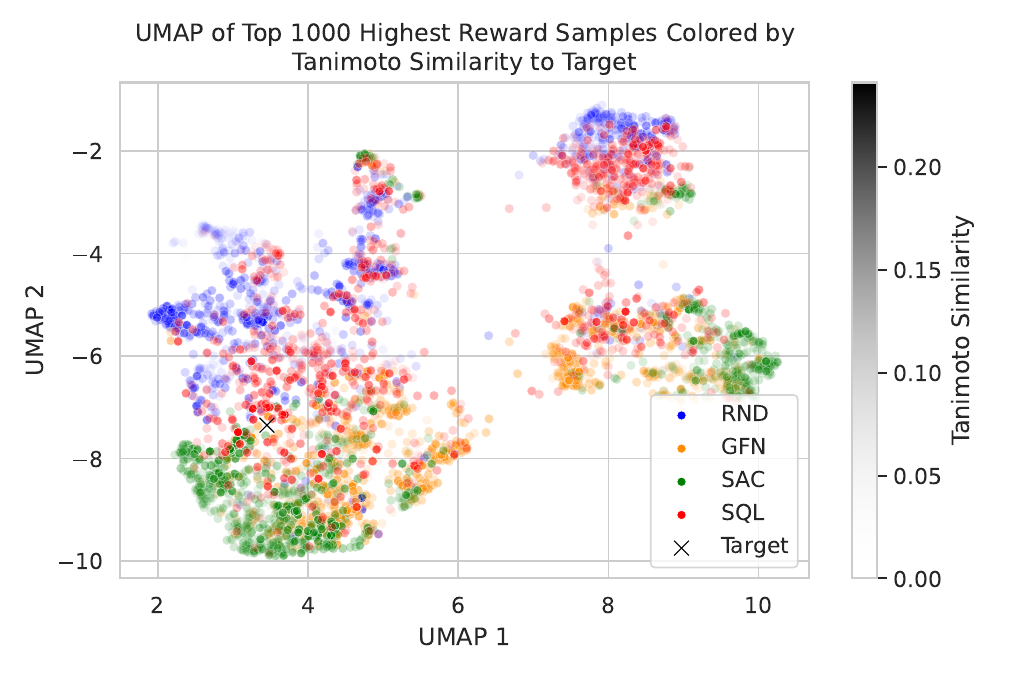}
\includegraphics[width=0.24\textwidth]{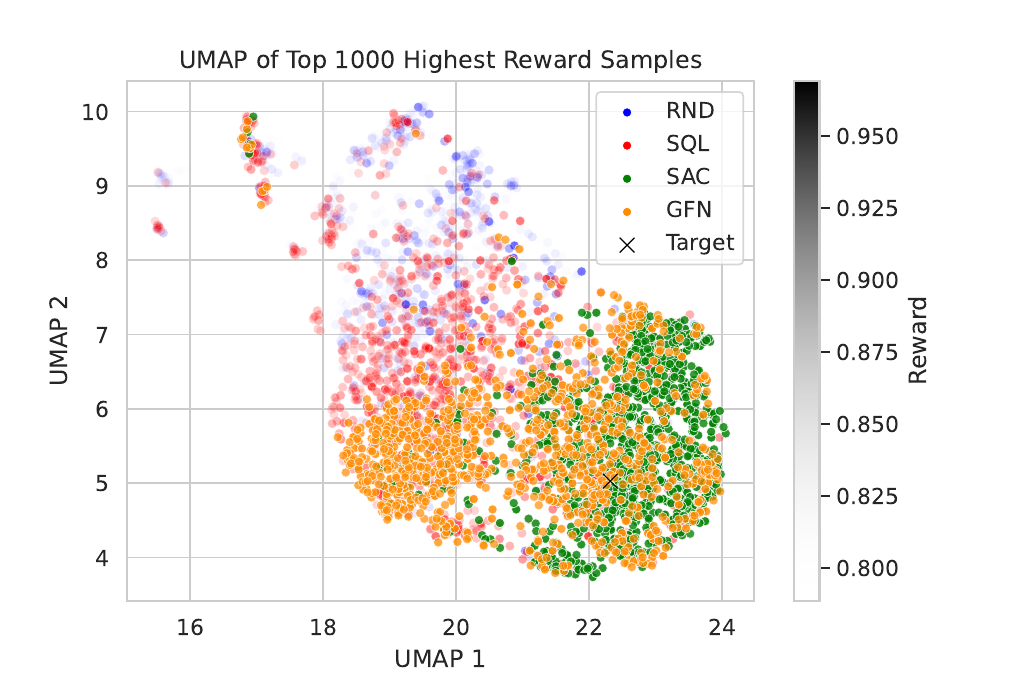}
\includegraphics[width=0.24\textwidth]{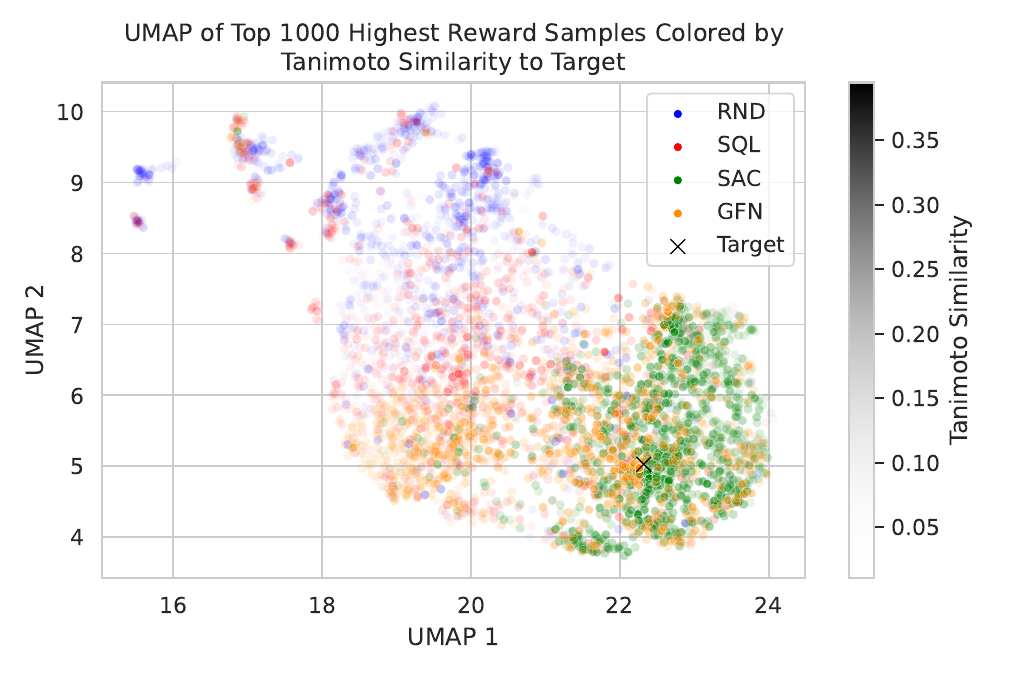}

\rotatebox{90}{\hspace{1.5em}\# 8505}
\includegraphics[width=0.24\textwidth]{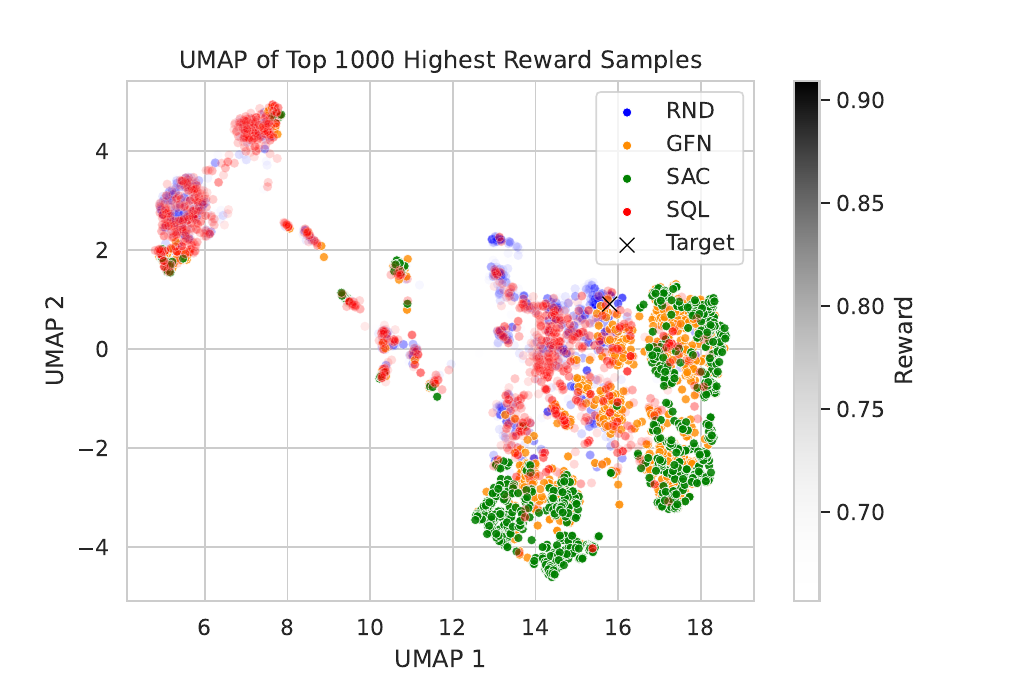}
\includegraphics[width=0.24\textwidth]{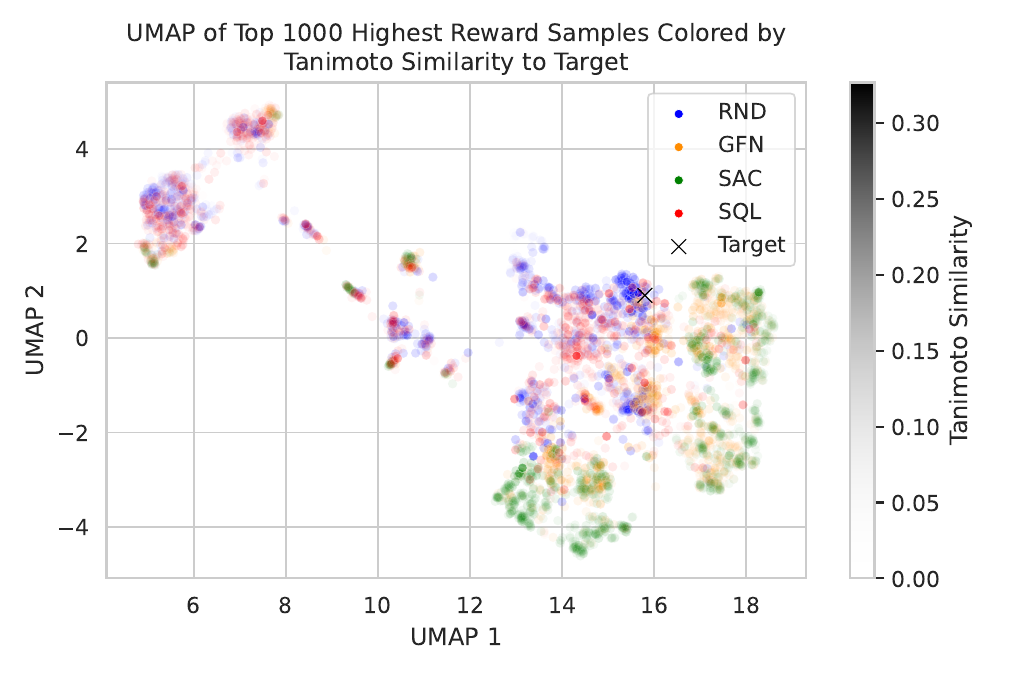}
\includegraphics[width=0.24\textwidth]{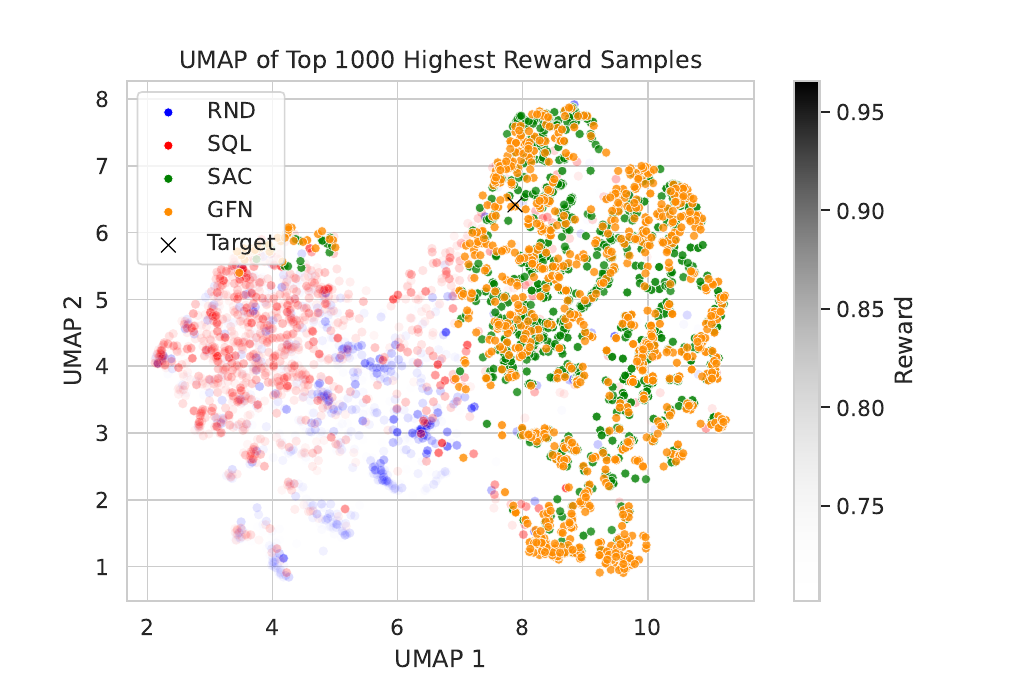}
\includegraphics[width=0.24\textwidth]{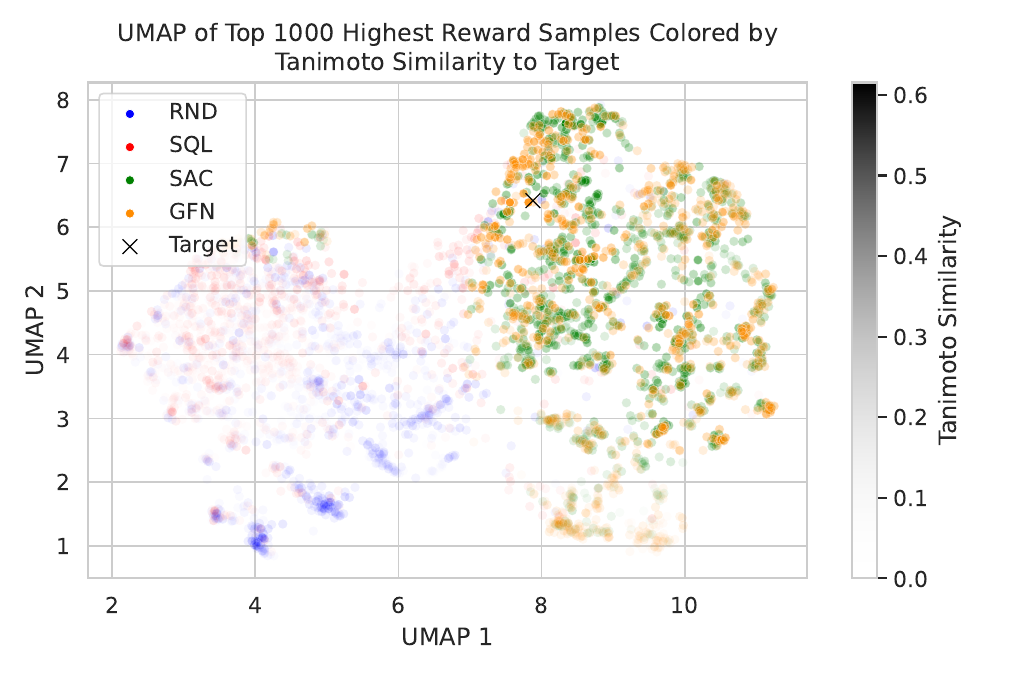}

\rotatebox{90}{\hspace{1.5em}\# 8636}
\includegraphics[width=0.24\textwidth]{res/umap/8636-all-morph/umap_rew.pdf}
\includegraphics[width=0.24\textwidth]{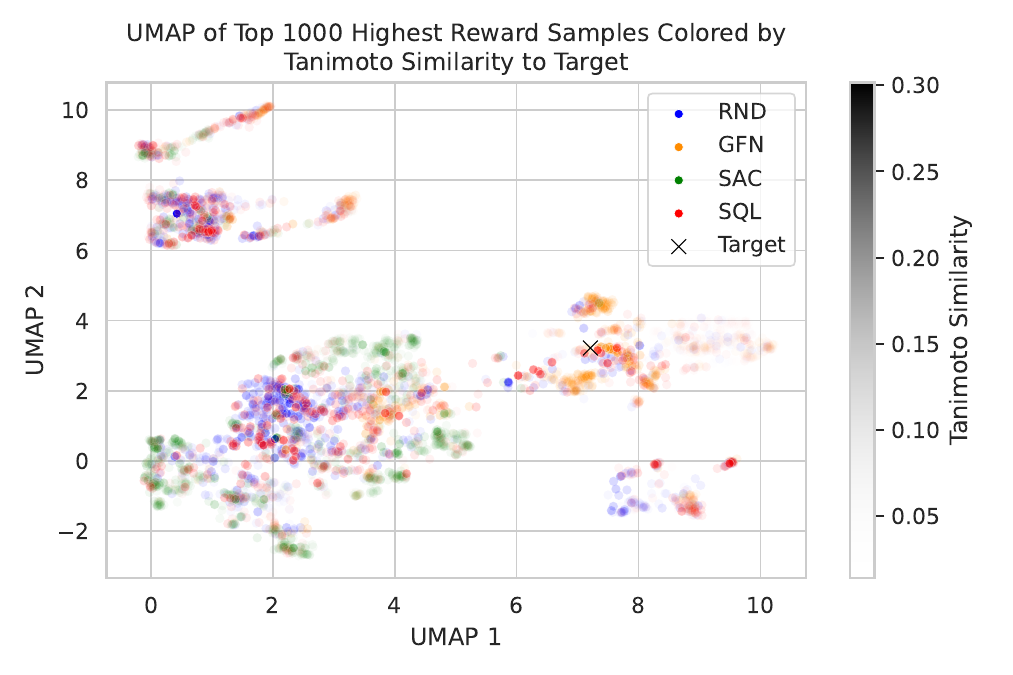}
\includegraphics[width=0.24\textwidth]{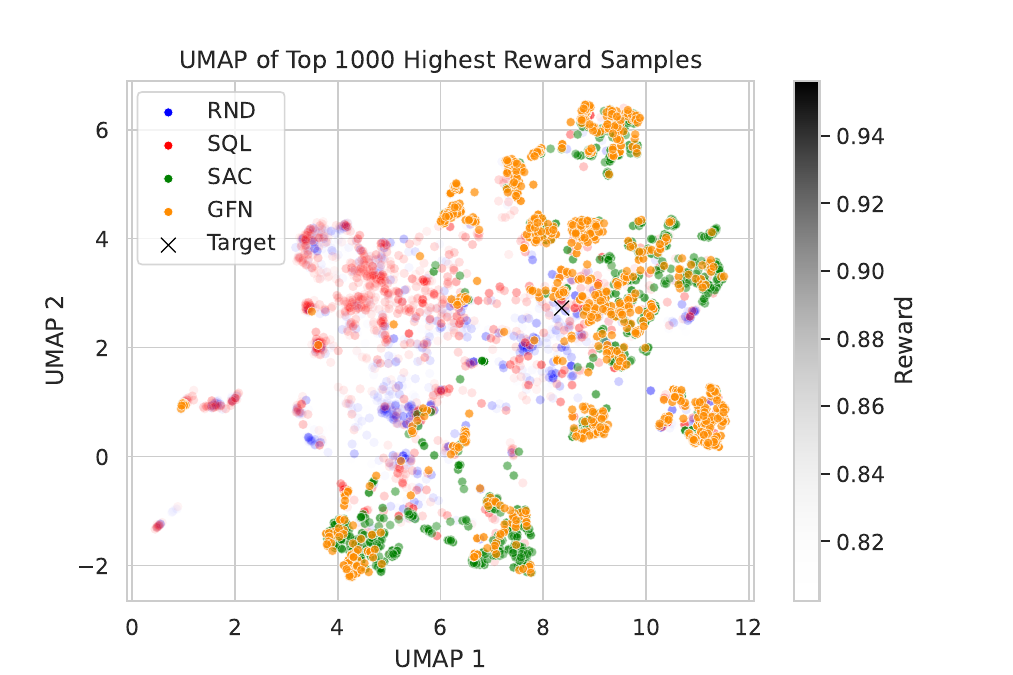}
\includegraphics[width=0.24\textwidth]{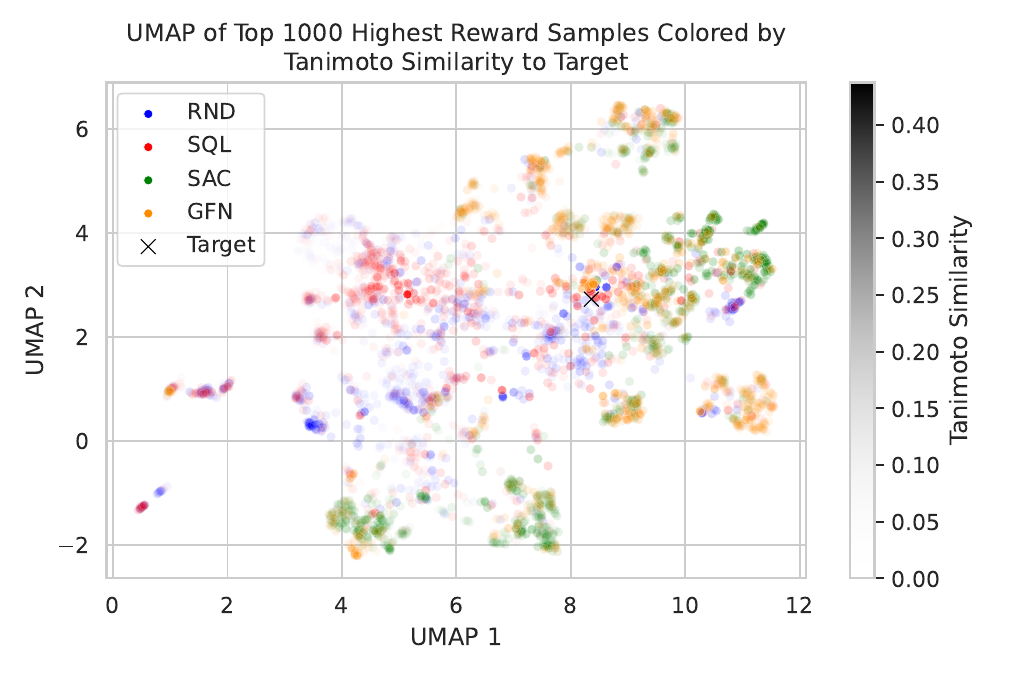}

\rotatebox{90}{\hspace{1.5em}\# 8949}
\includegraphics[width=0.24\textwidth]{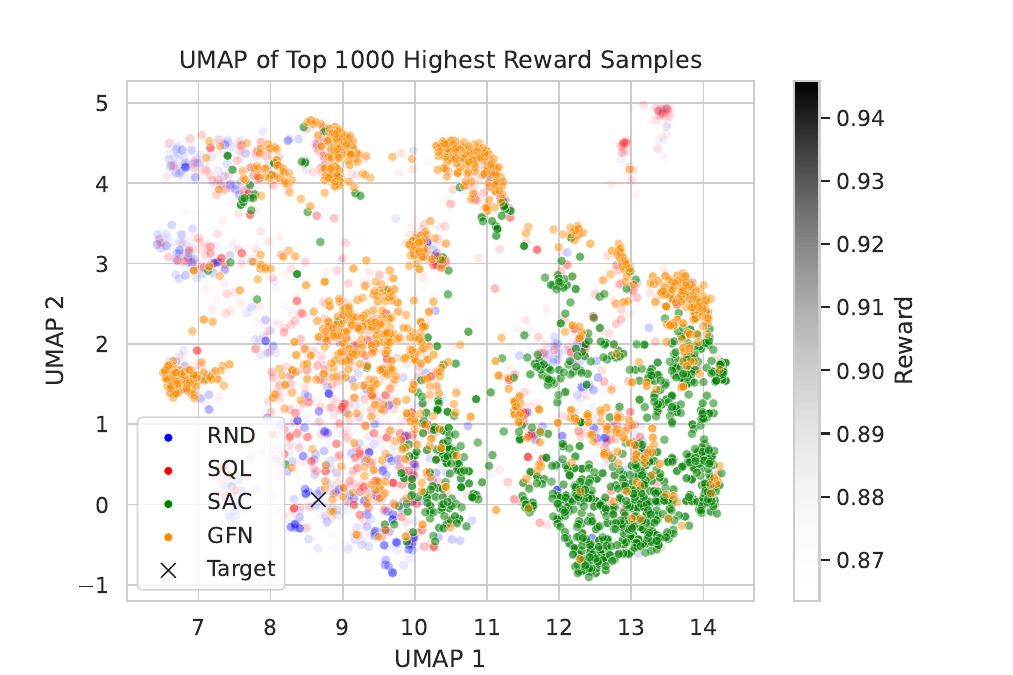}
\includegraphics[width=0.24\textwidth]{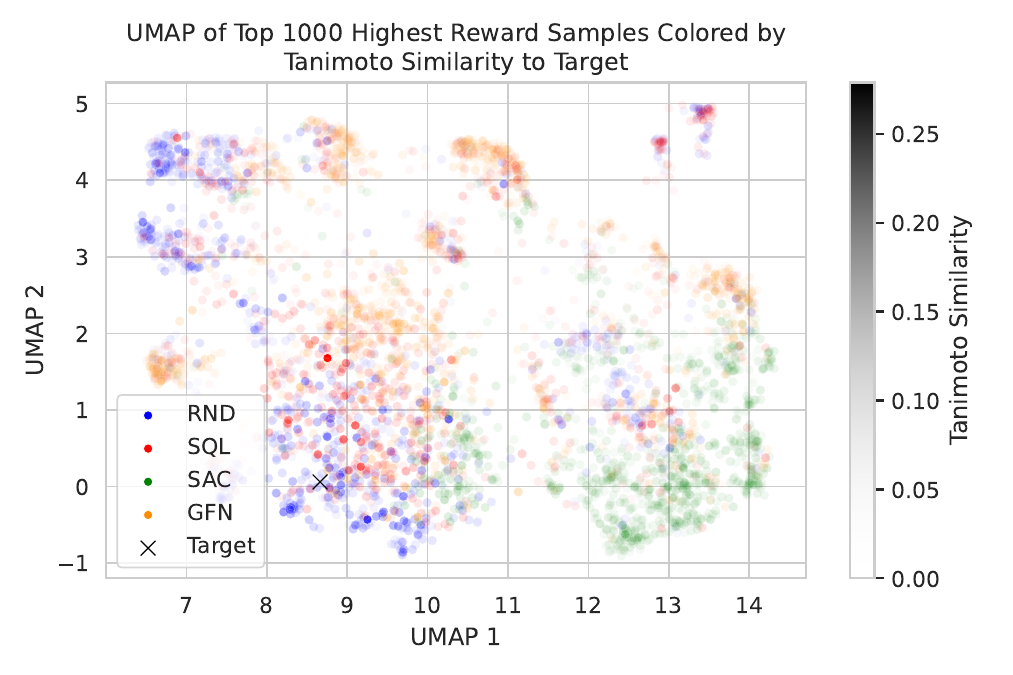}
\includegraphics[width=0.24\textwidth]{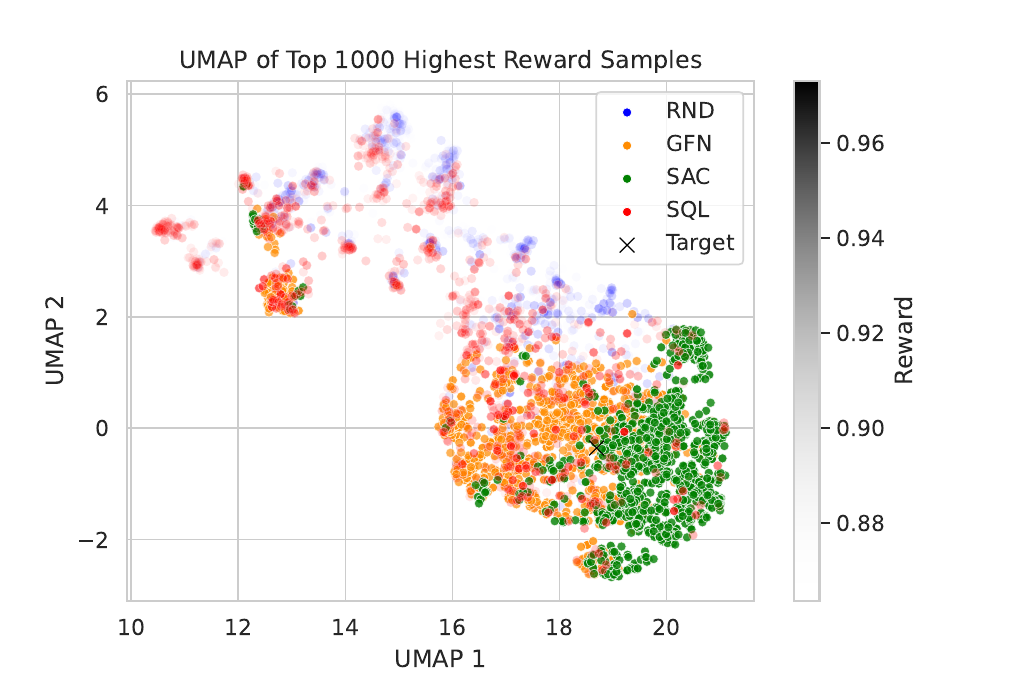}
\includegraphics[width=0.24\textwidth]{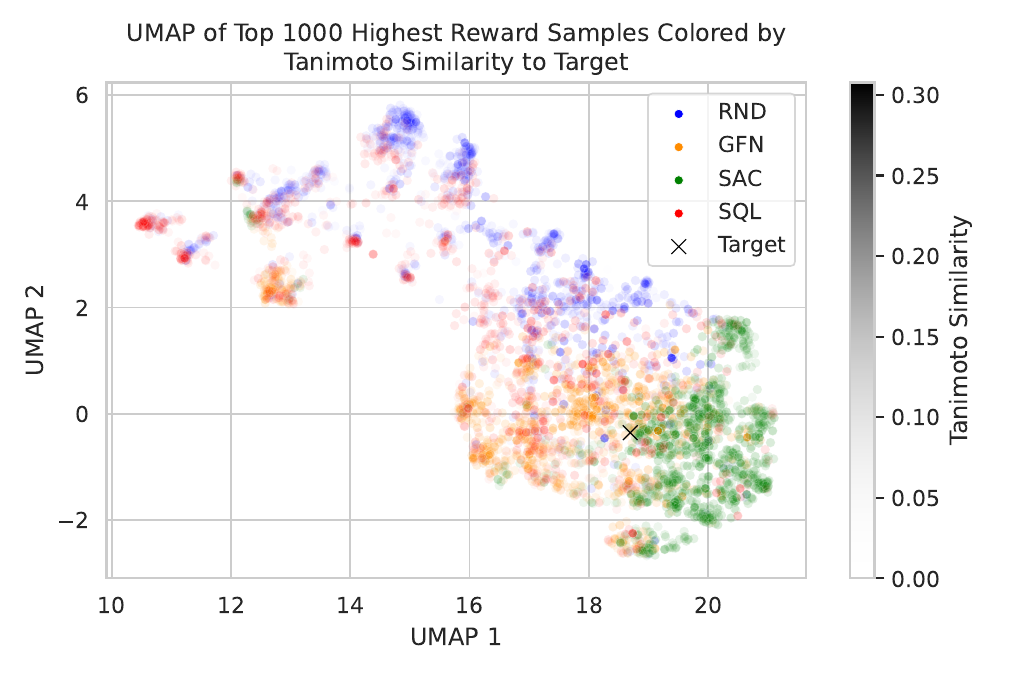}

\rotatebox{90}{\hspace{1.5em}\# 9300}
\includegraphics[width=0.24\textwidth]{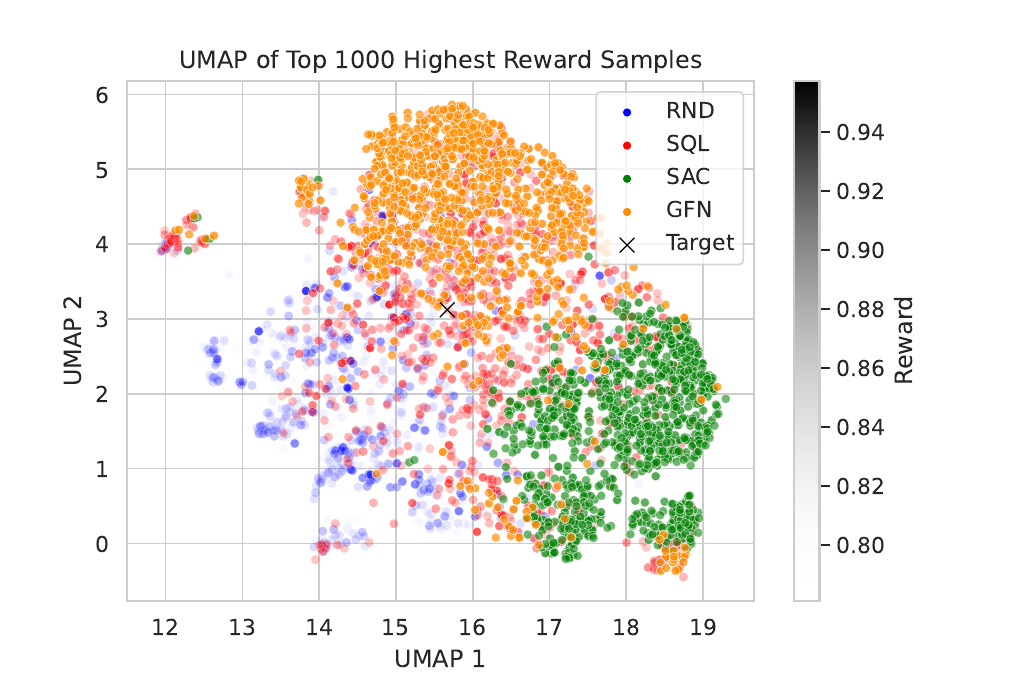}
\includegraphics[width=0.24\textwidth]{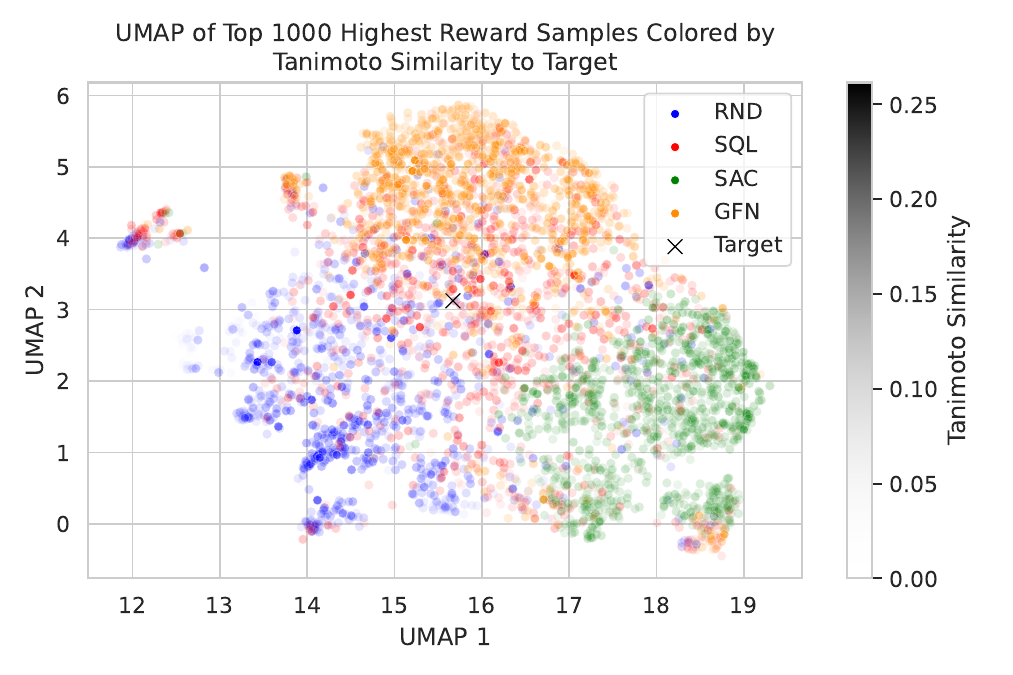}
\includegraphics[width=0.24\textwidth]{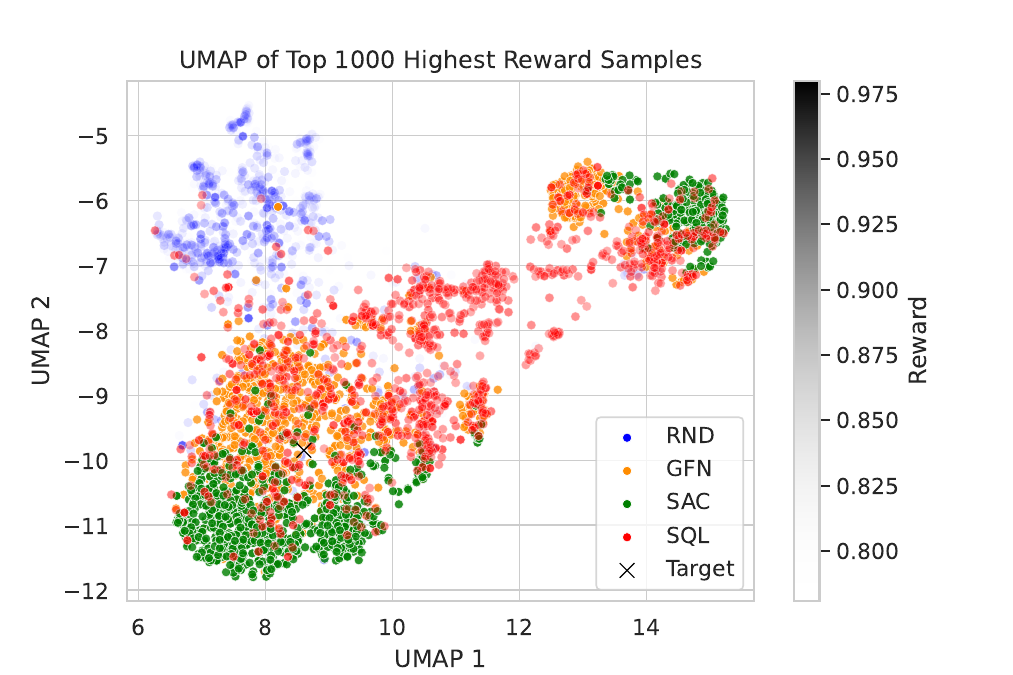}
\includegraphics[width=0.24\textwidth]{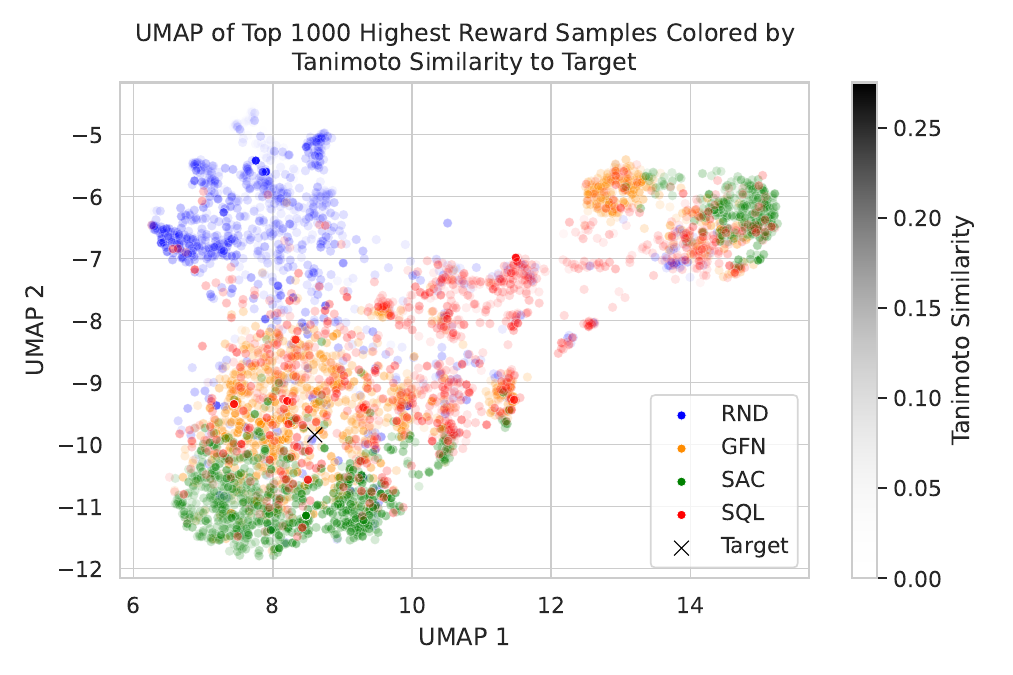}

\rotatebox{90}{\hspace{1.5em}\# 9445}
\includegraphics[width=0.24\textwidth]{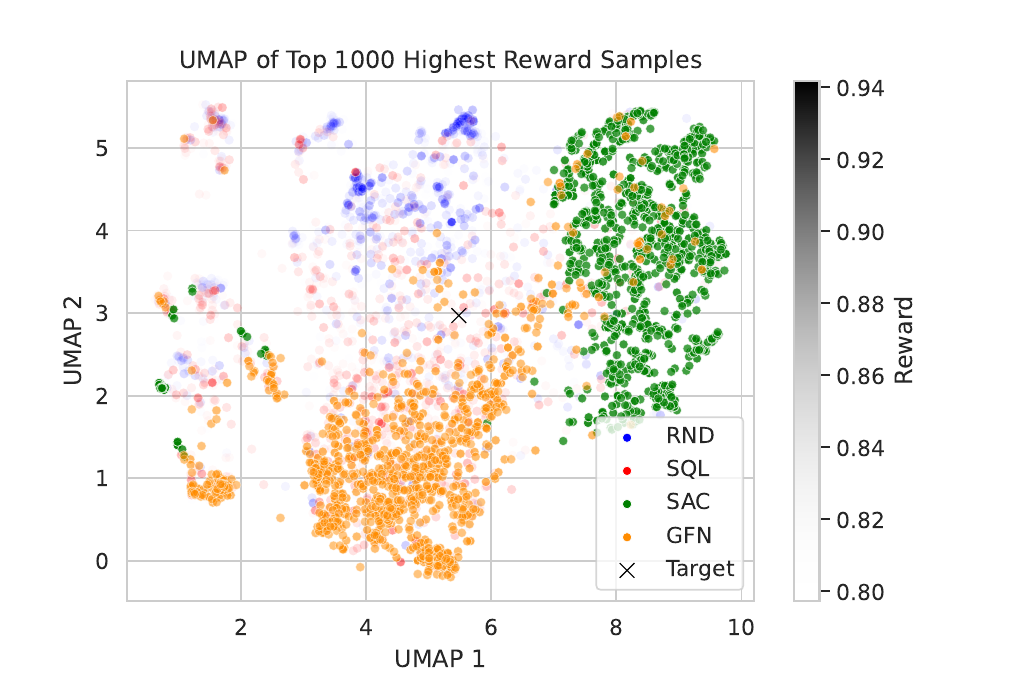}
\includegraphics[width=0.24\textwidth]{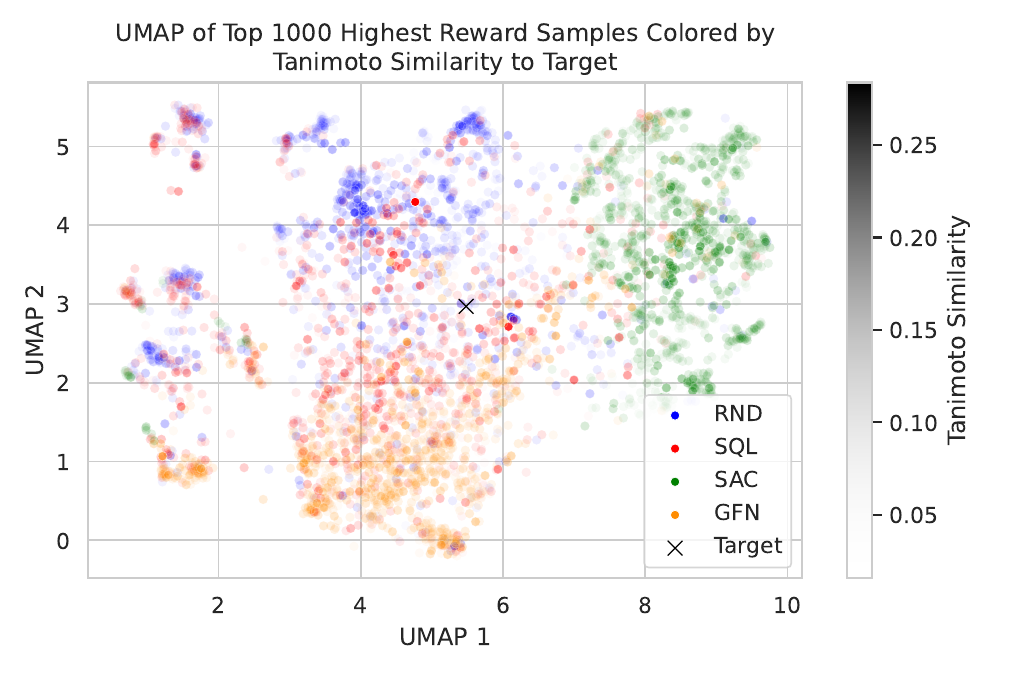}
\includegraphics[width=0.24\textwidth]{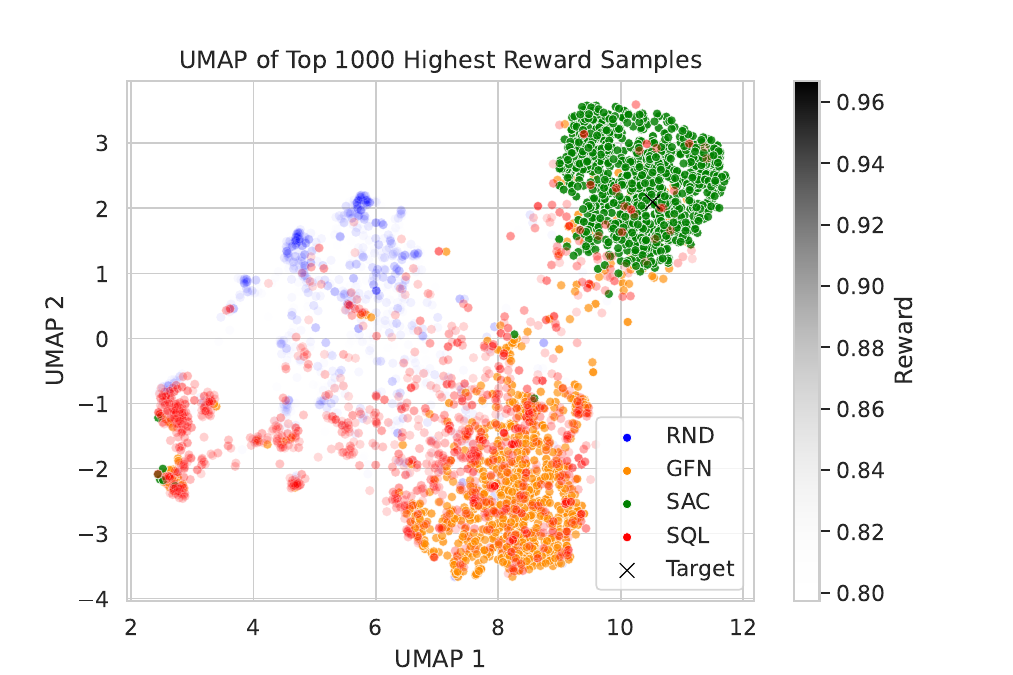}
\includegraphics[width=0.24\textwidth]{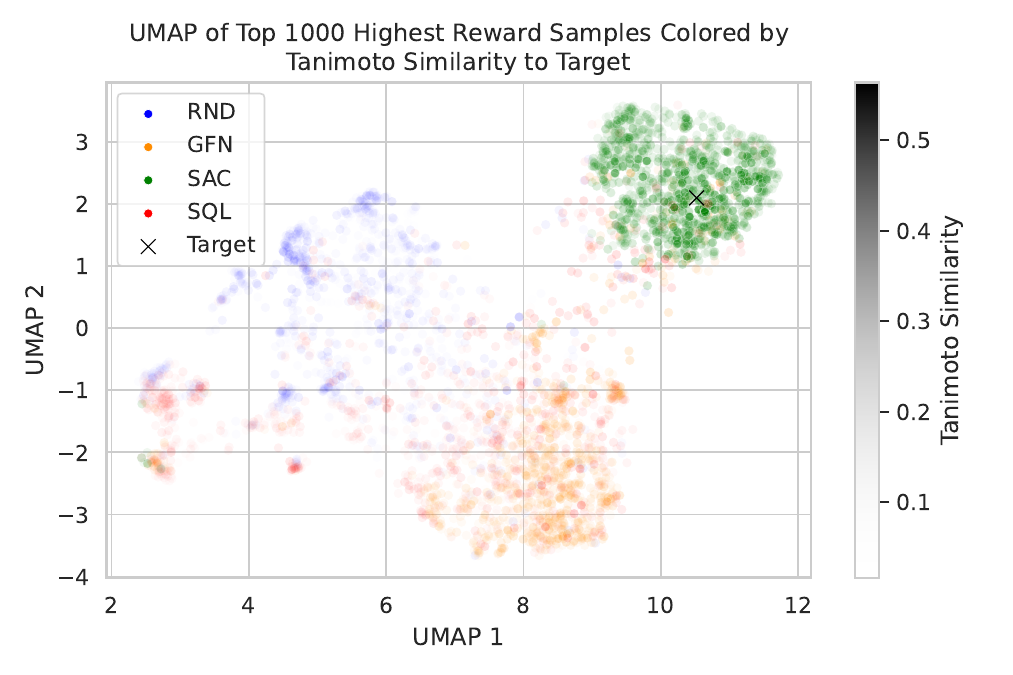}

\rotatebox{90}{\hspace{1.5em}\# 9476}
\includegraphics[width=0.24\textwidth]{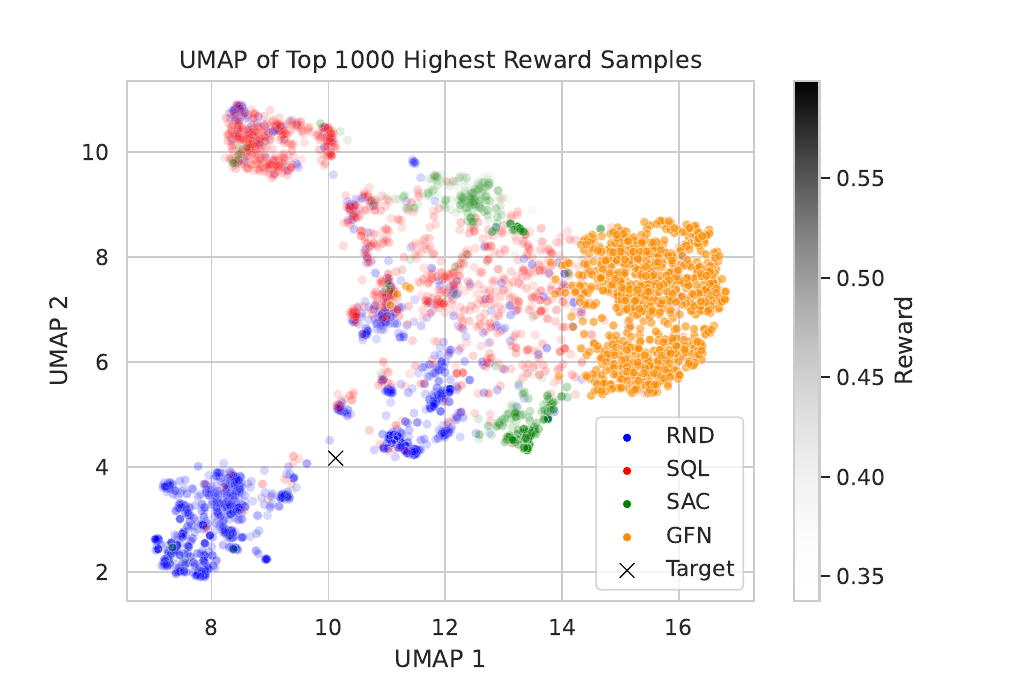}
\includegraphics[width=0.24\textwidth]{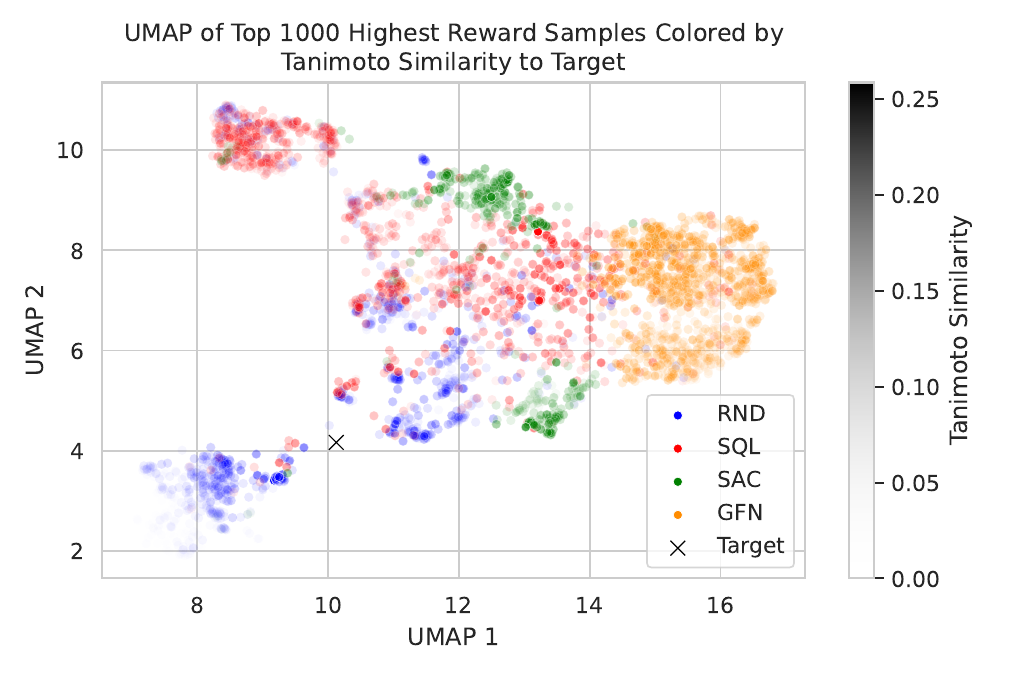}
\includegraphics[width=0.24\textwidth]{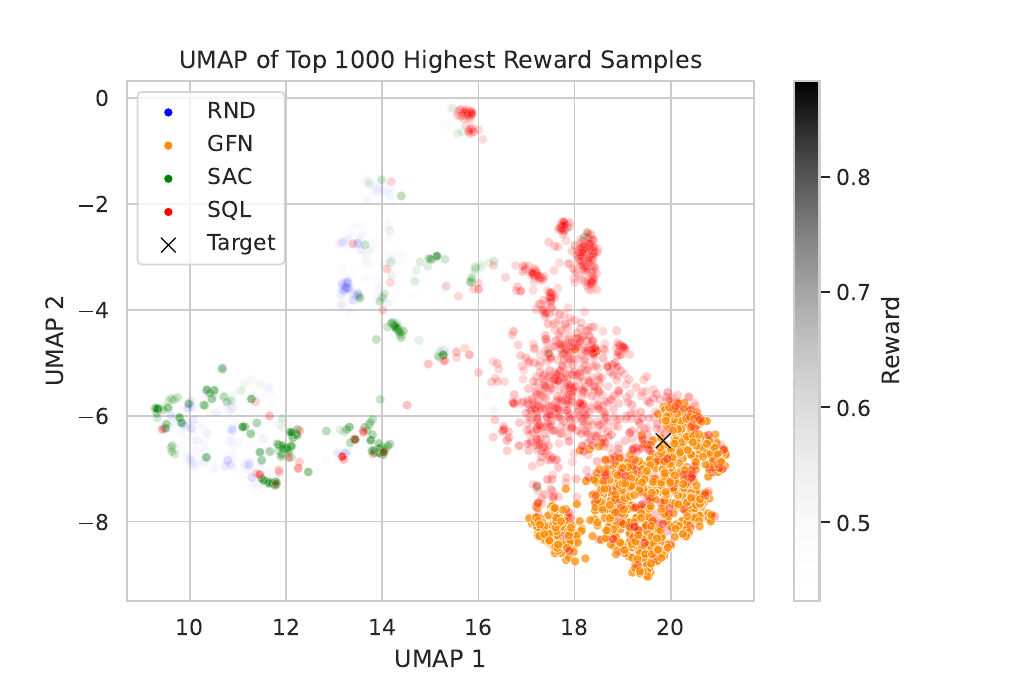}
\includegraphics[width=0.24\textwidth]{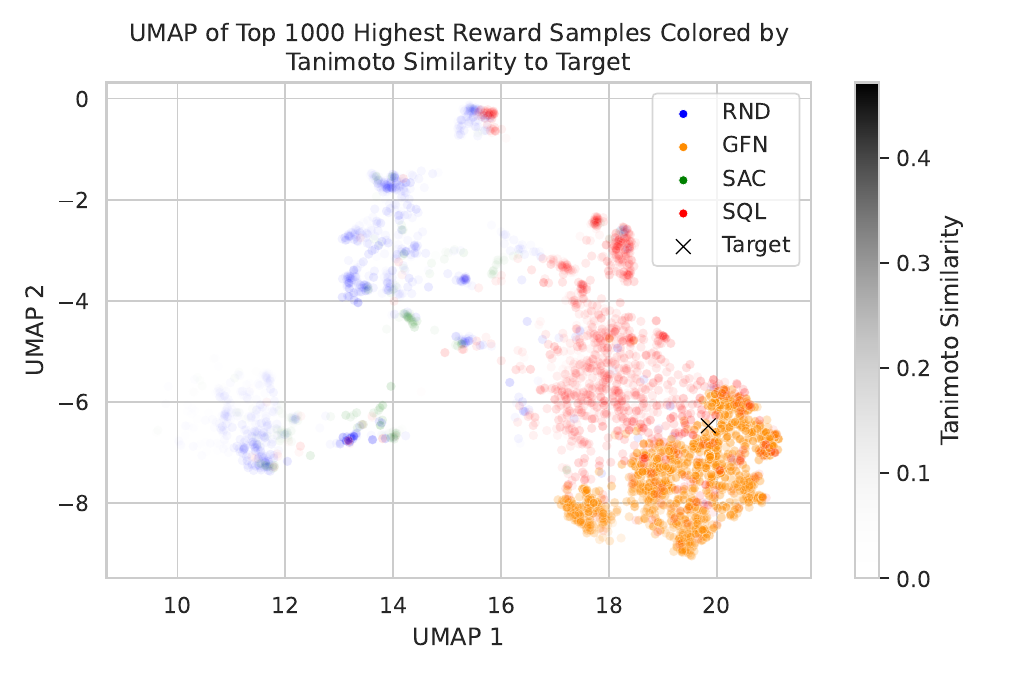}

\rotatebox{90}{\hspace{1.25em}\# 10075}
\includegraphics[width=0.24\textwidth]{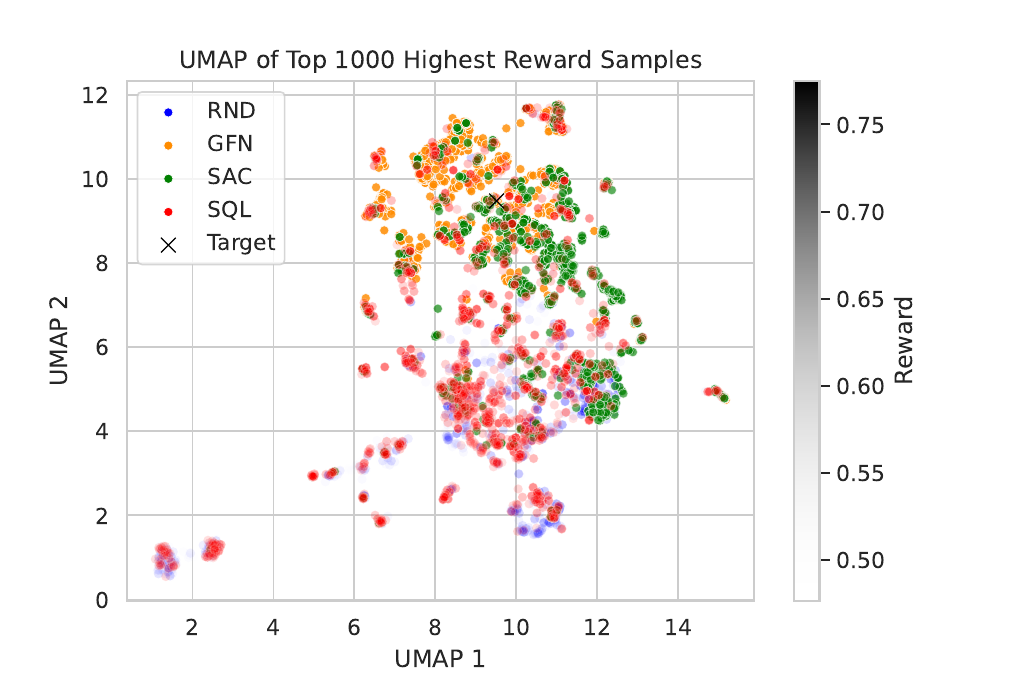}
\includegraphics[width=0.24\textwidth]{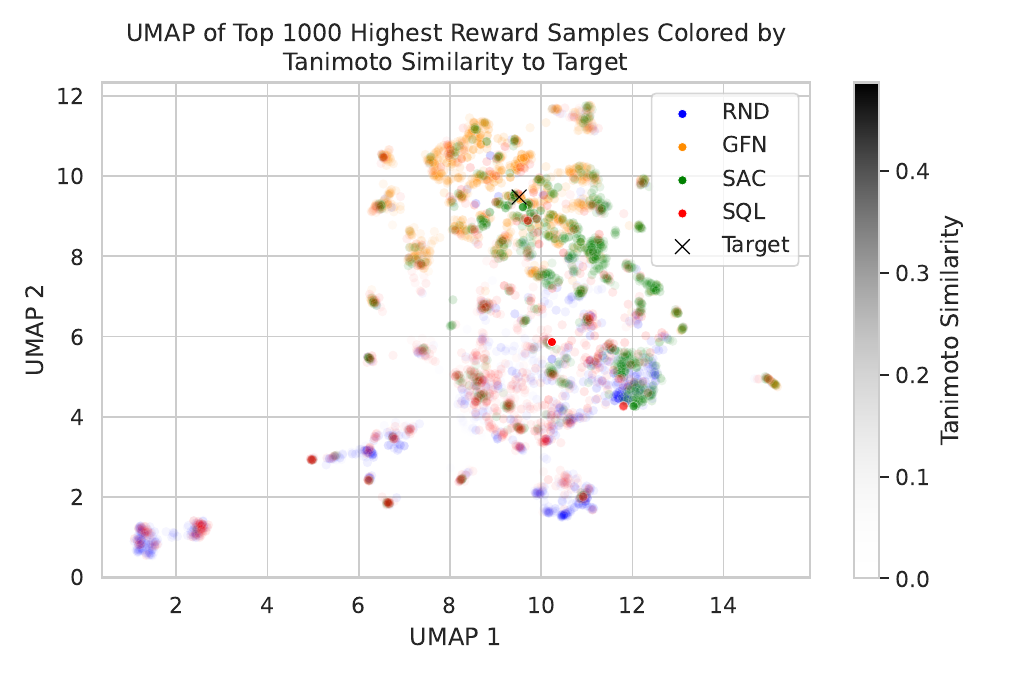}
\includegraphics[width=0.24\textwidth]{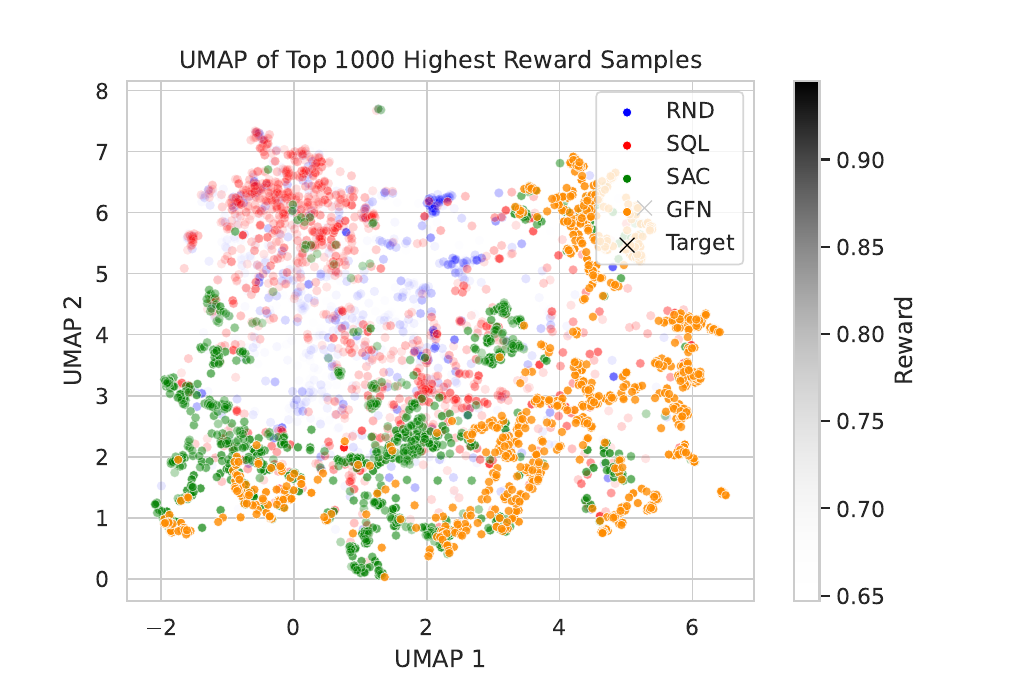}
\includegraphics[width=0.24\textwidth]{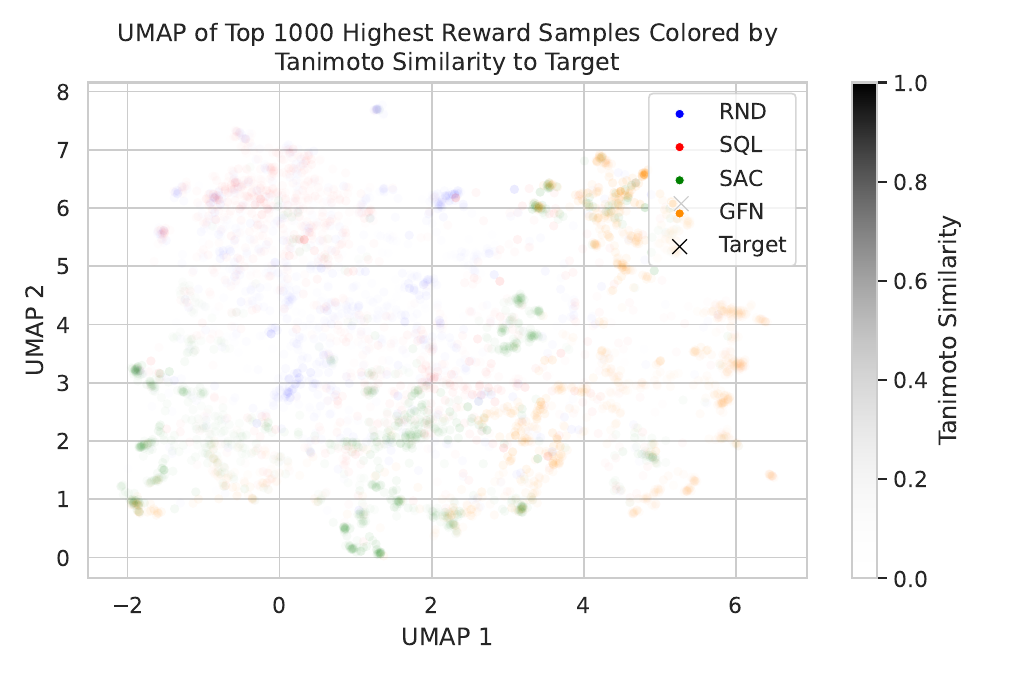}

\rotatebox{90}{\hspace{1.25em}\# 10816}
\includegraphics[width=0.24\textwidth]{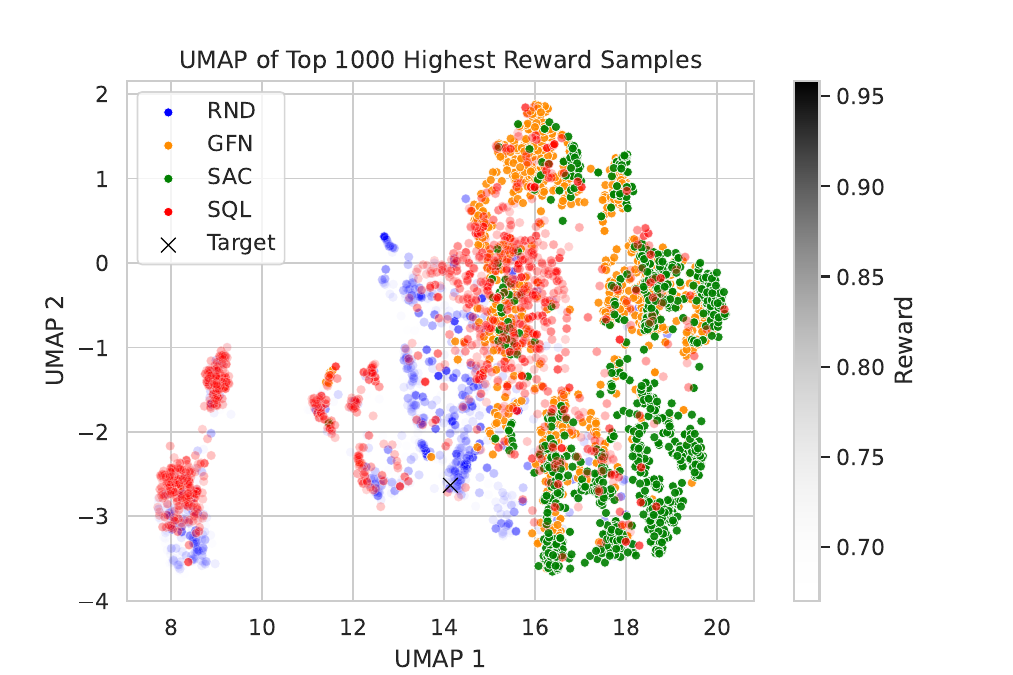}
\includegraphics[width=0.24\textwidth]{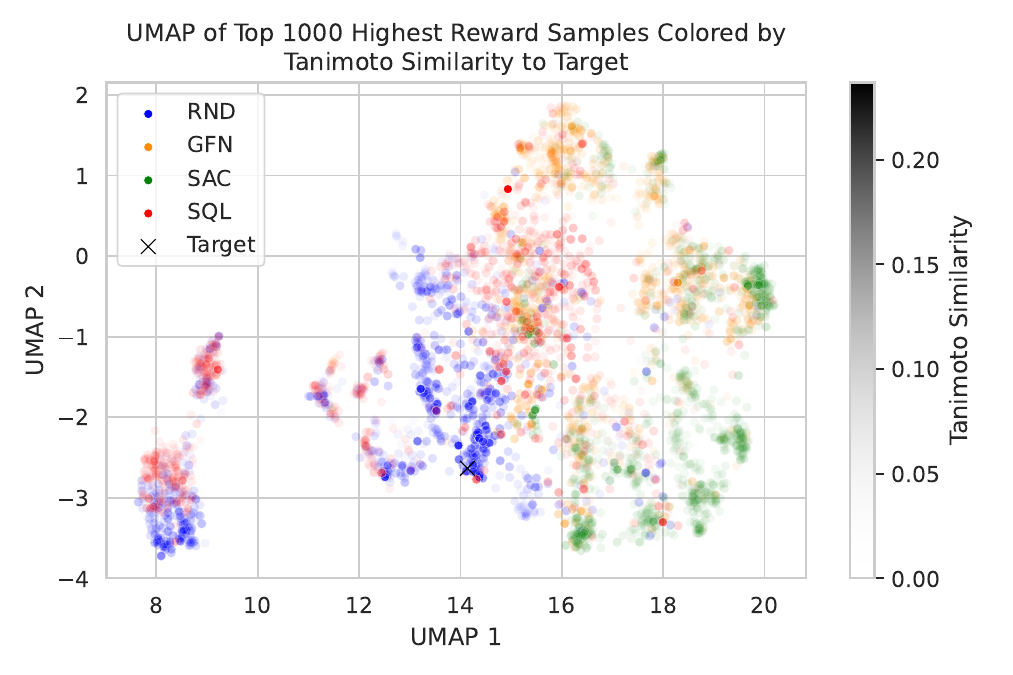}
\includegraphics[width=0.24\textwidth]{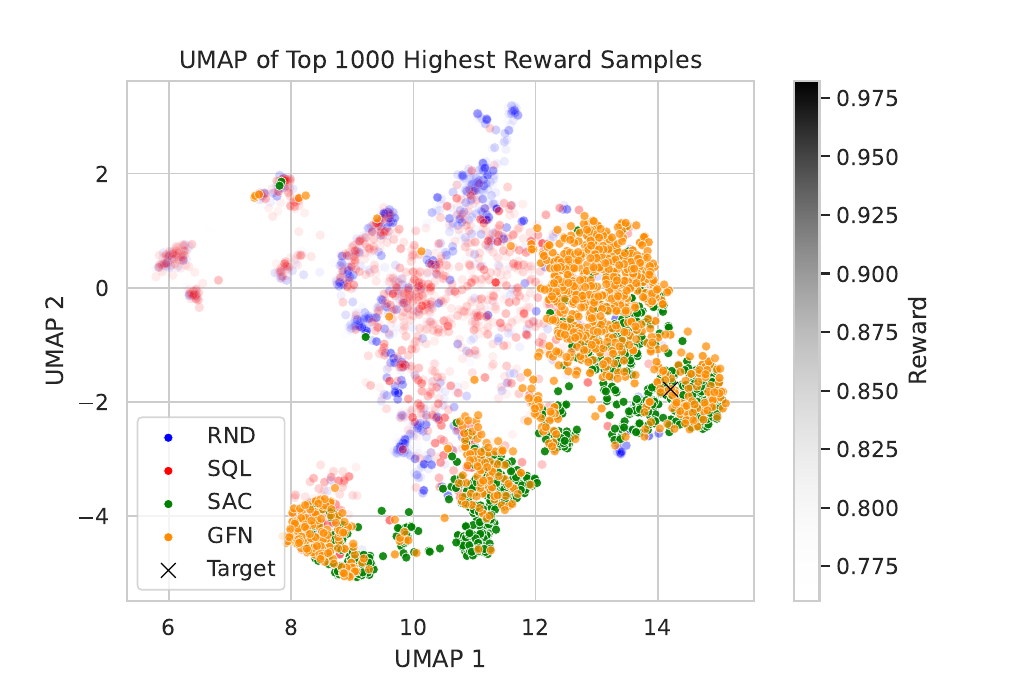}
\includegraphics[width=0.24\textwidth]{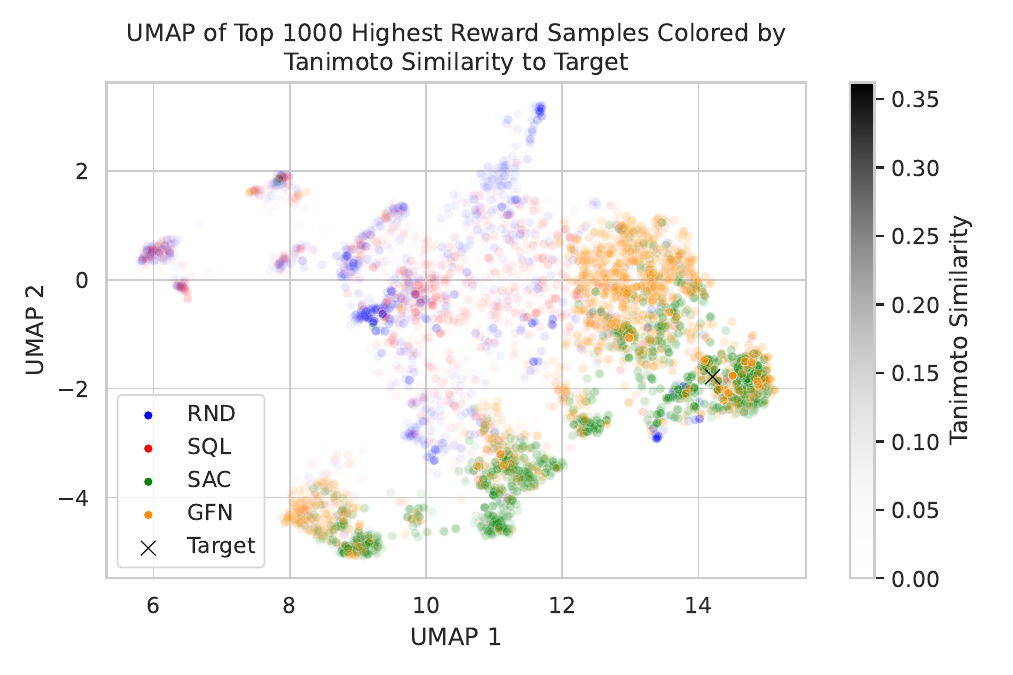}

\rotatebox{90}{\hspace{1.25em}\# 12071}
\includegraphics[width=0.24\textwidth]{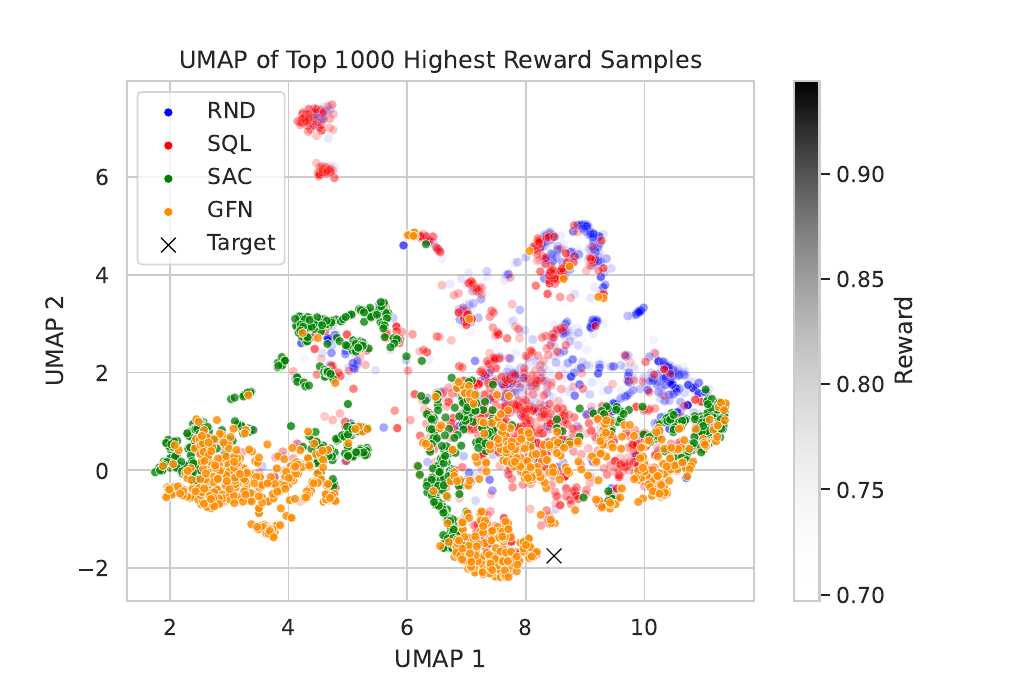}
\includegraphics[width=0.24\textwidth]{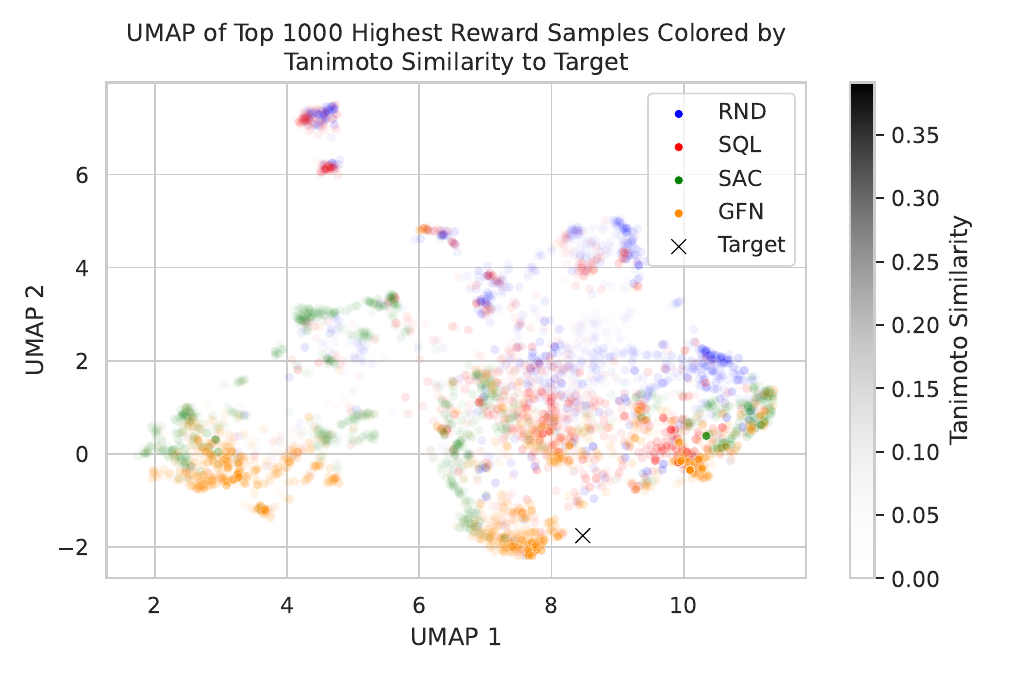}
\includegraphics[width=0.24\textwidth]{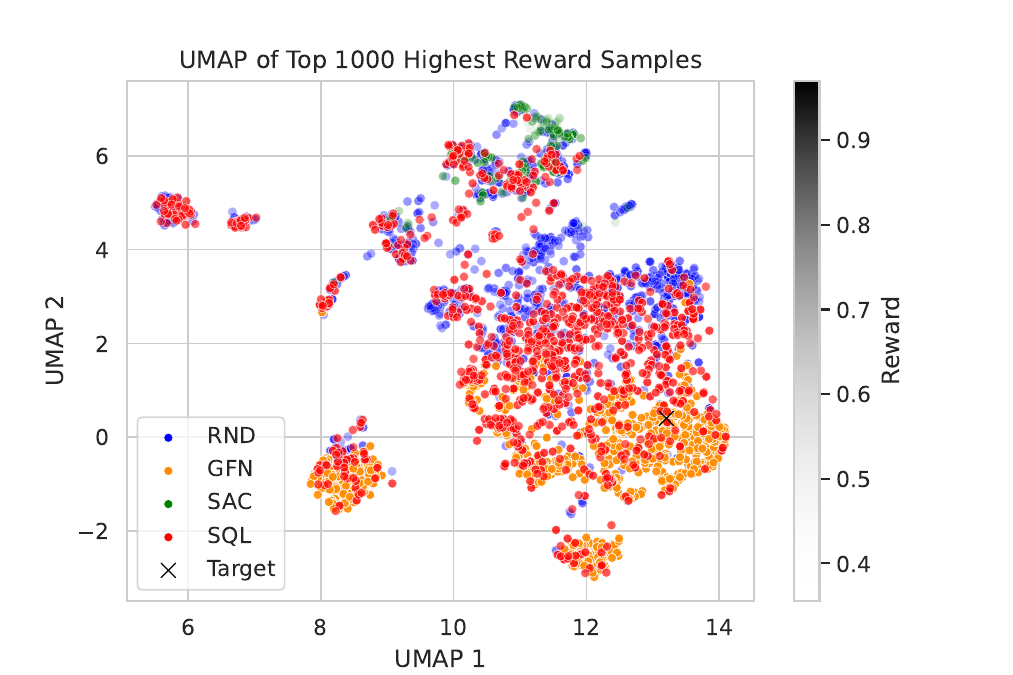}
\includegraphics[width=0.24\textwidth]{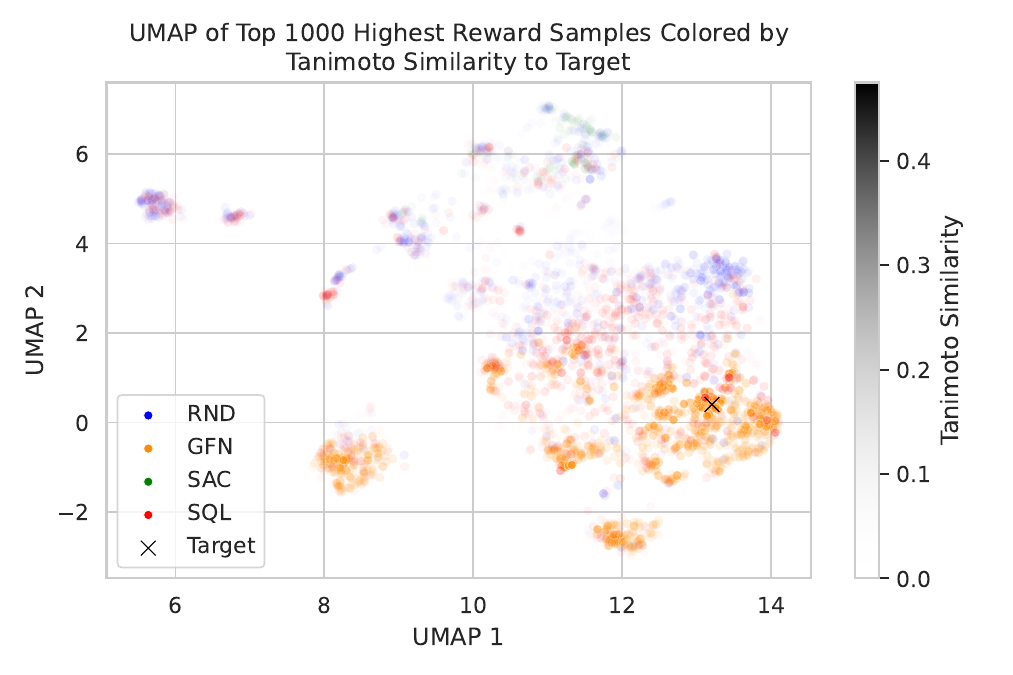}

\rotatebox{90}{\hspace{1.25em}\# 12662}
\includegraphics[width=0.24\textwidth]{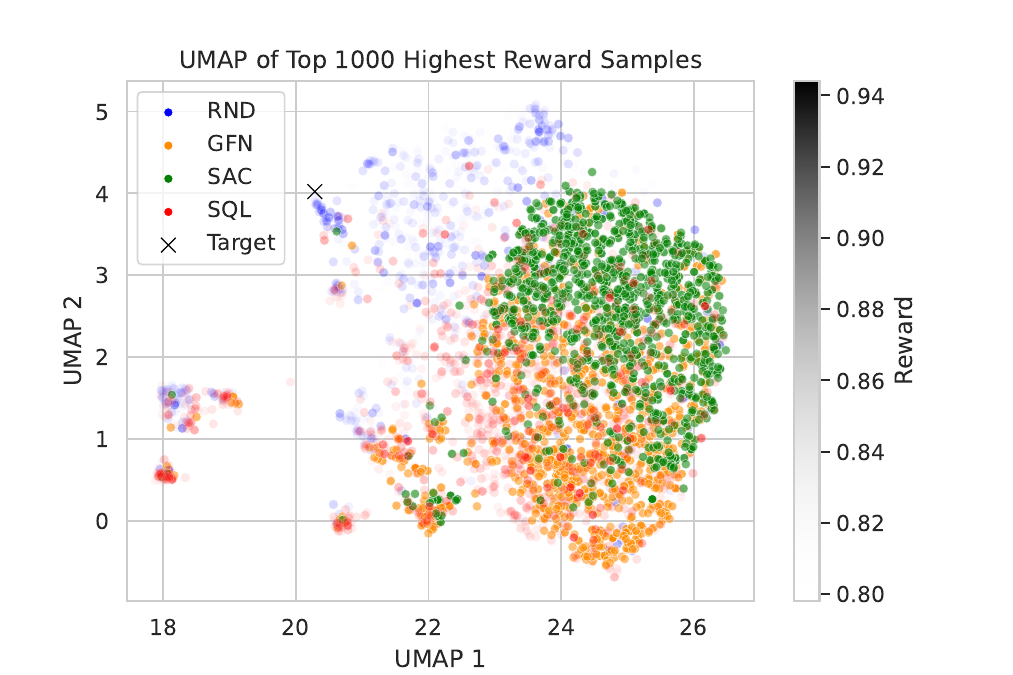}
\includegraphics[width=0.24\textwidth]{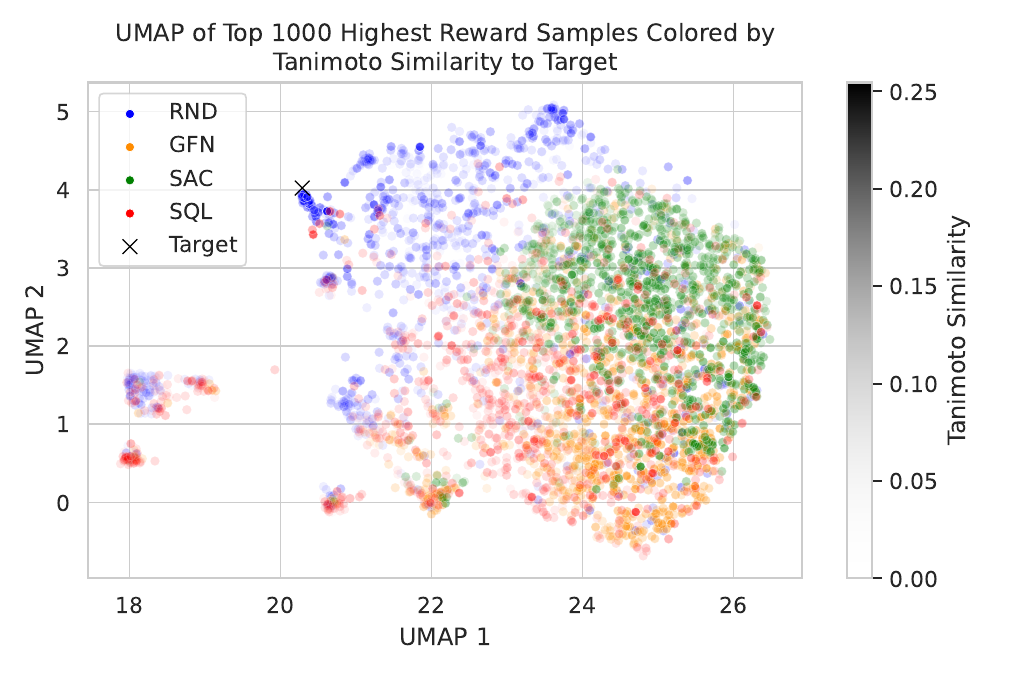}
\includegraphics[width=0.24\textwidth]{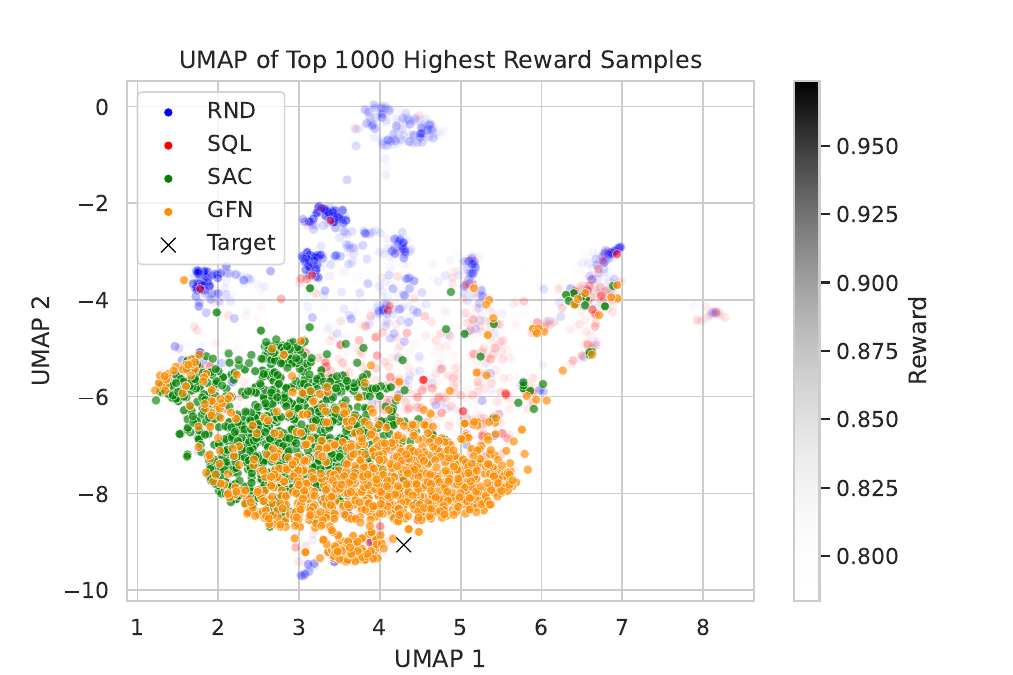}
\includegraphics[width=0.24\textwidth]{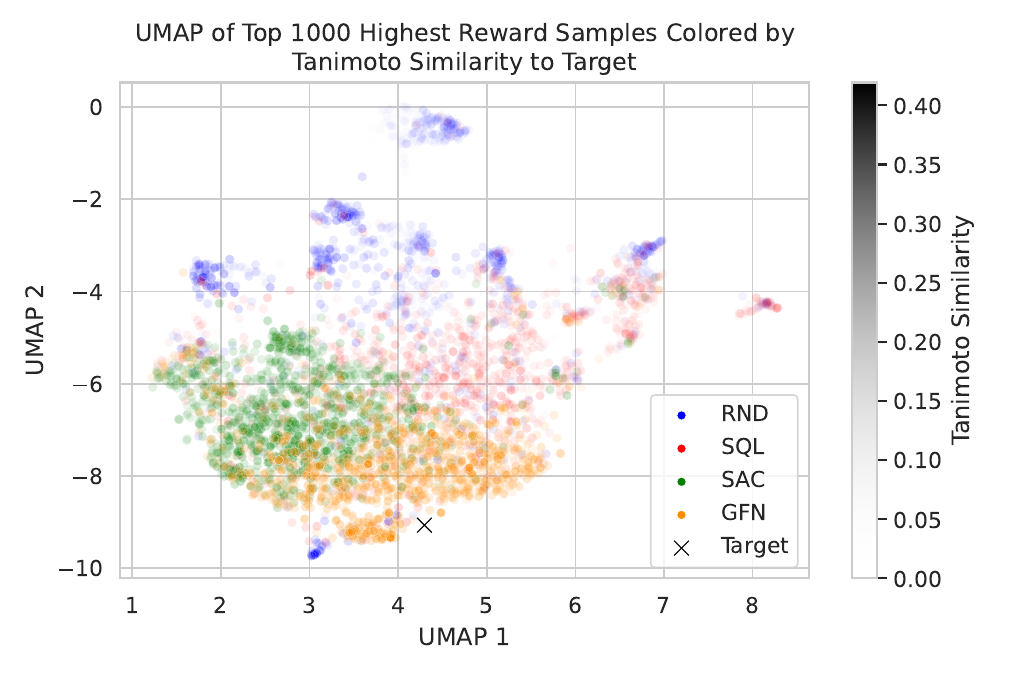}

\rotatebox{90}{\hspace{1.25em}\# 15575}
\includegraphics[width=0.24\textwidth]{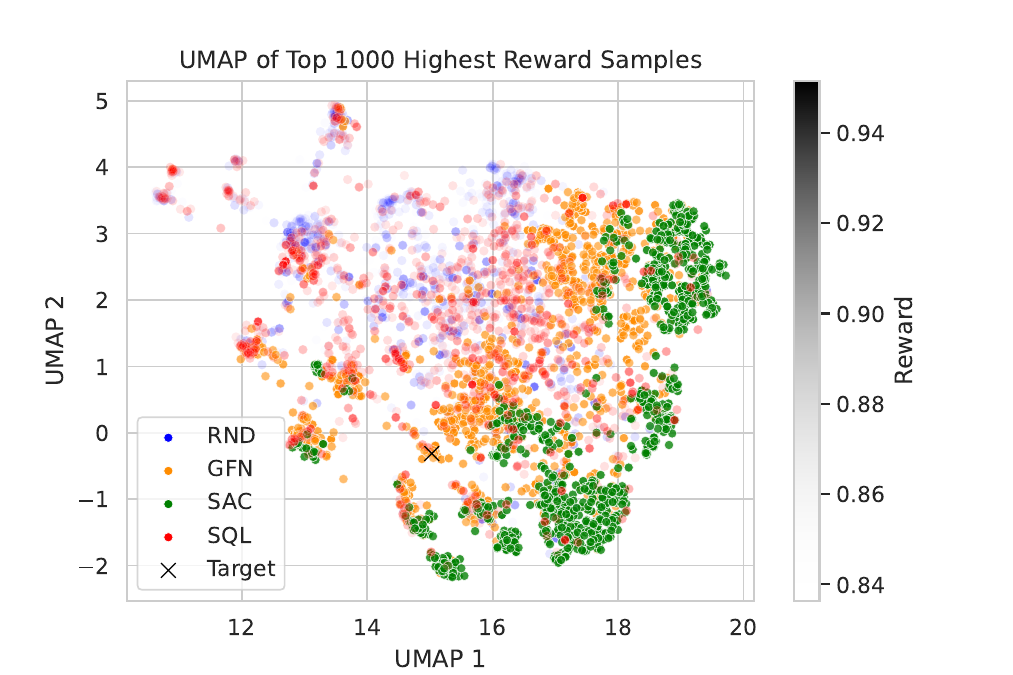}
\includegraphics[width=0.24\textwidth]{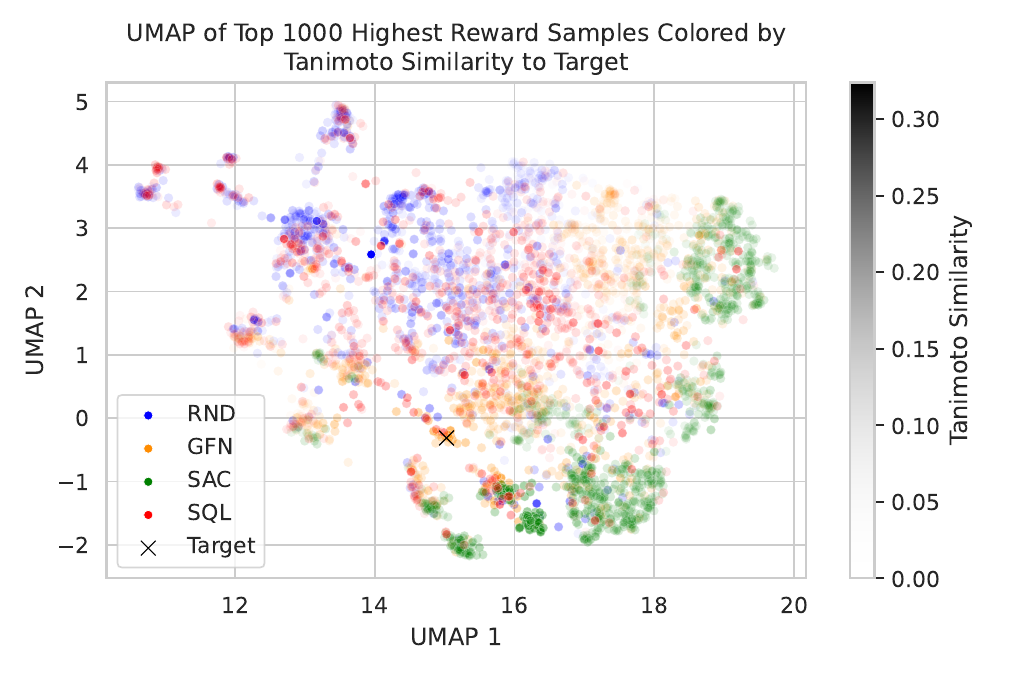}
\includegraphics[width=0.24\textwidth]{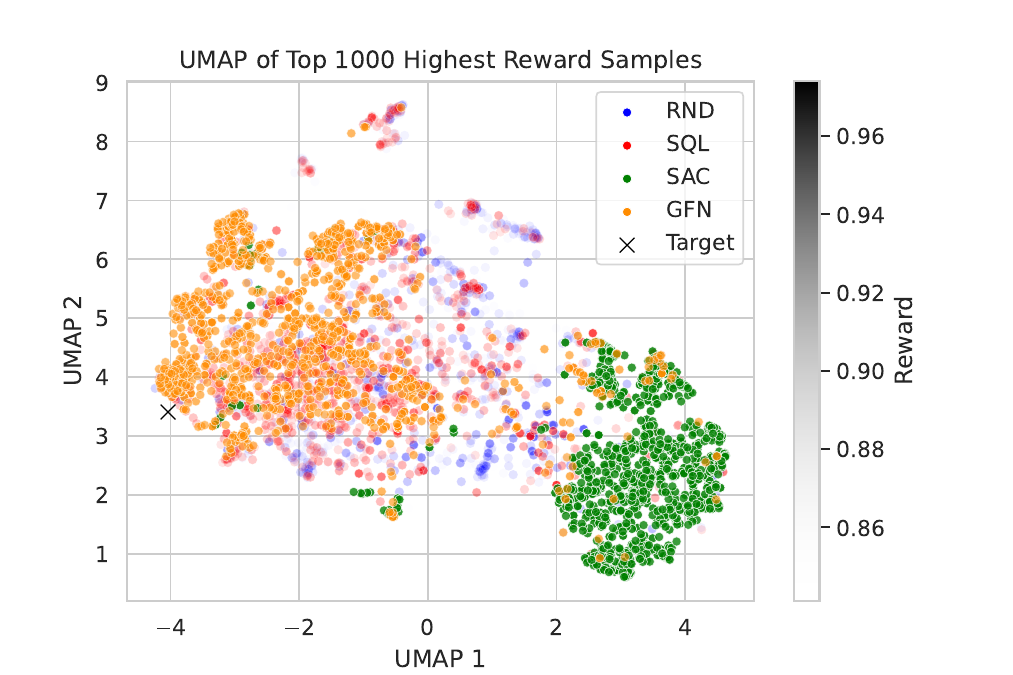}
\includegraphics[width=0.24\textwidth]{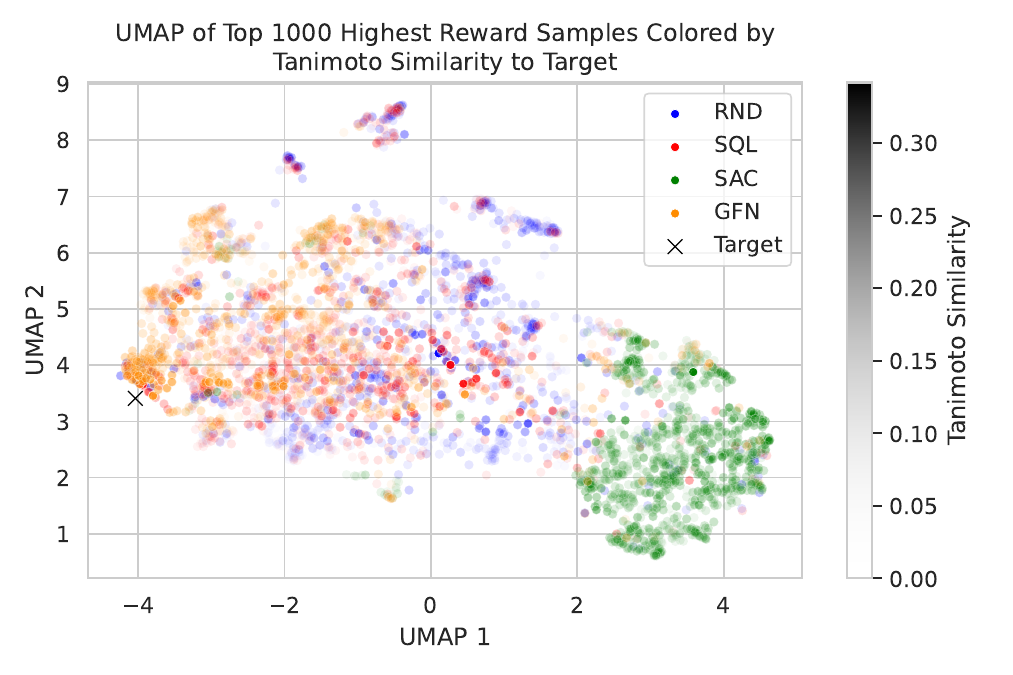}

\subsection{High reward samples}
\begin{figure}[!htb]
    \centering
    \includegraphics[width=0.49\textwidth]{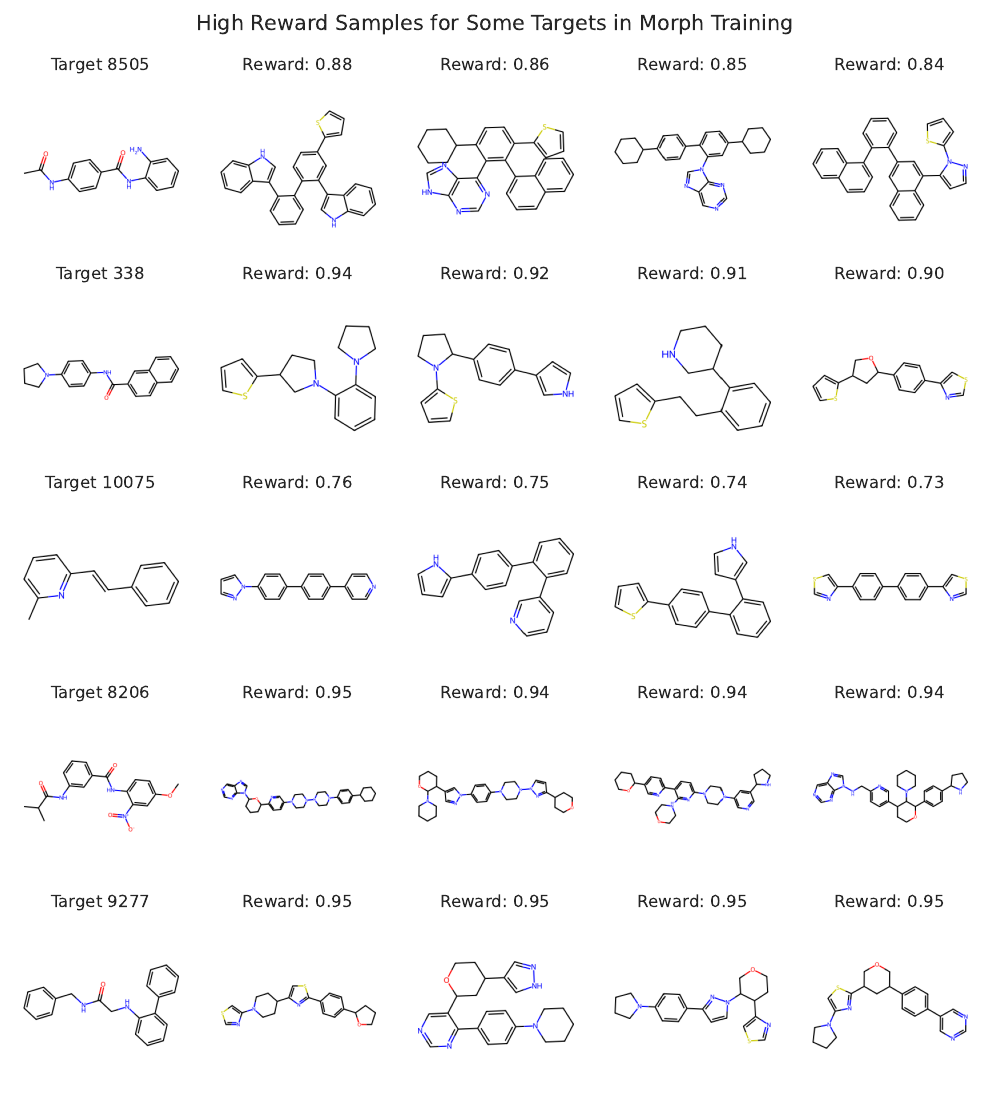}
    \includegraphics[width=0.49\textwidth]{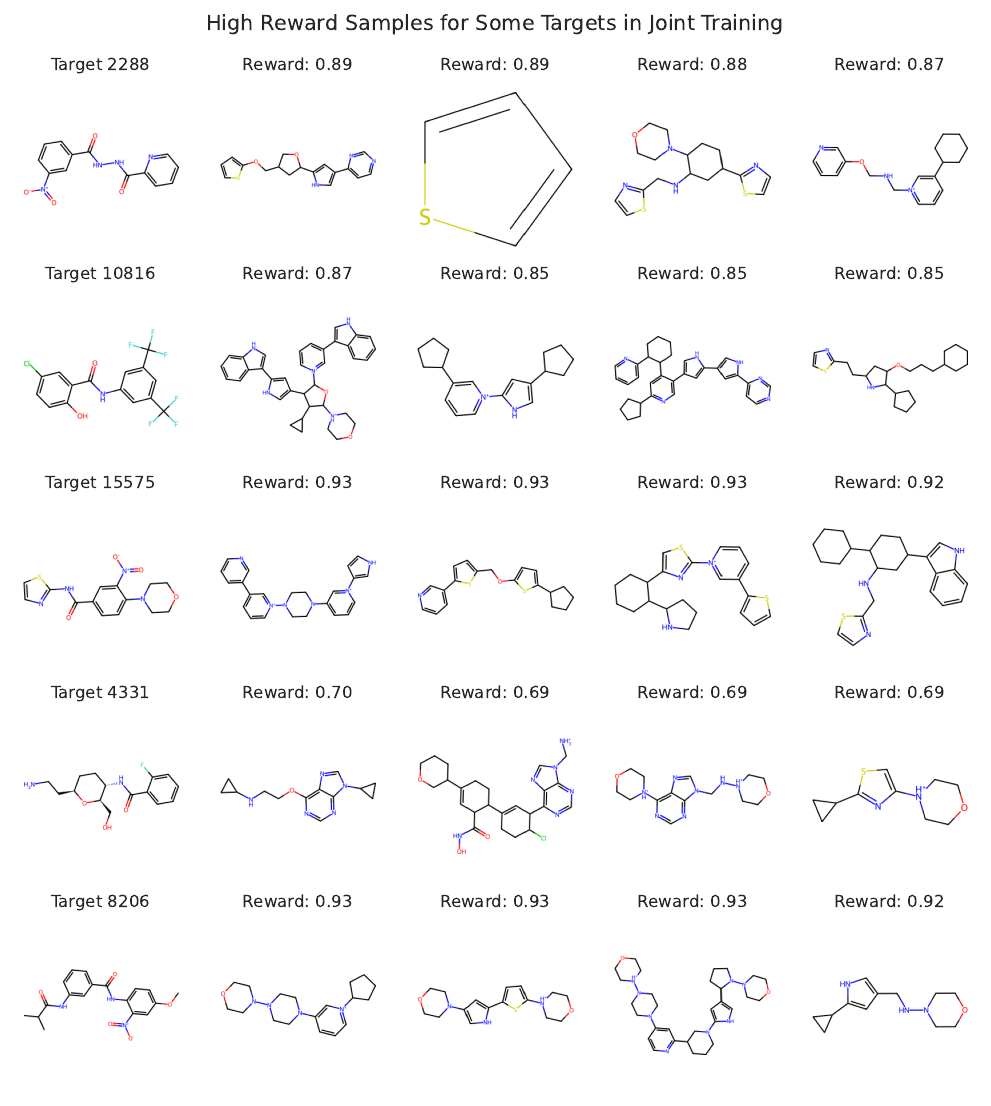}
    \caption{High reward samples from GFlowNets trained against morphology-only targets (left) and joint morphology and structure guided targets (right).}
    \label{apx:high_rew_samples}
\end{figure}

In \Cref{apx:high_rew_samples} we plot a set of randomly selected high reward samples from GFlowNets trained on targets in a) morphology-only setting (left) and b) joint morphology and structure-guided setting (right).

\subsection{GMC cross-modality alignment}
\label{apx:gmc_single}
We plot the pairwise cosine similarity between embeddings of all test set samples for each modality pairing of the GMC representation space. A high correlation indicates that the model learns to correctly align the different input domains, such that embeddings of associated modalities are aligned closer than distinct ones. As expected, GMC achieves a higher correlation when the joint modality is included in one of the axes since the joint latent space integrates signals from both the structure and morphology features, which is redundant with the other axis.

\begin{figure}[h!]
    \centering
    \includegraphics[width=\textwidth]{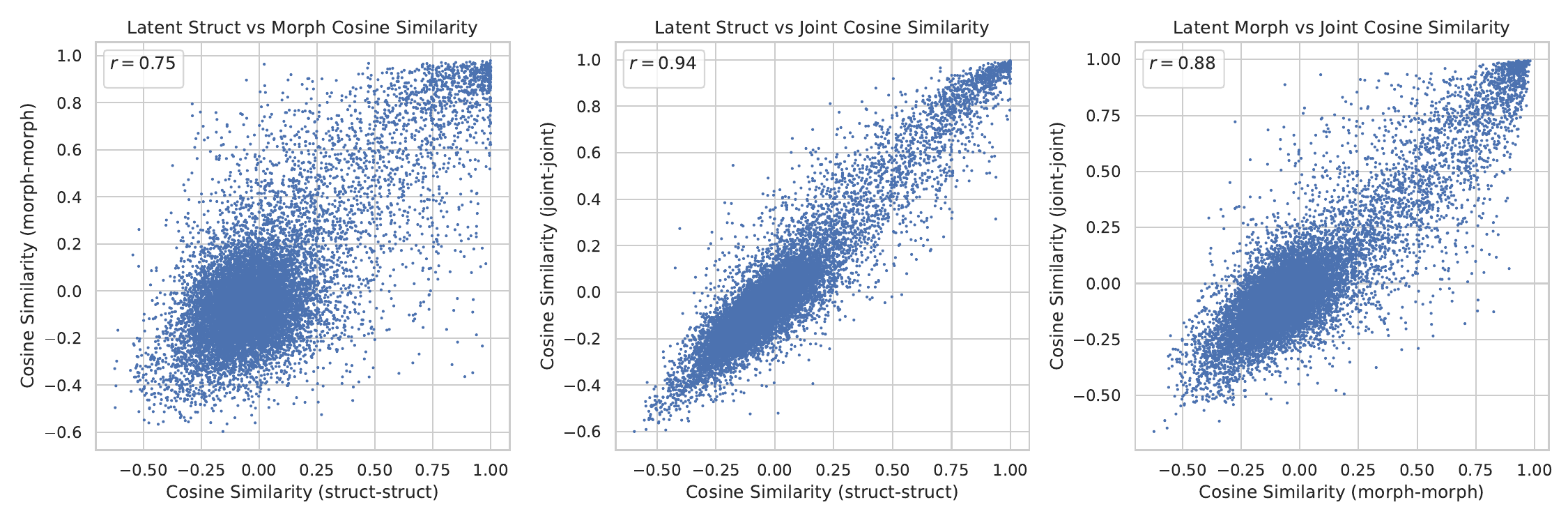}
    \caption{Pairwise cosine similarity of representations between structure and morphology domains (left), structure and joint domains (center), and morphology and joint domains (right). For each pair of samples in the test set, we plot the cosine similarity between their embeddings in a first modality (x-axis) and a second modality (y-axis).}
\end{figure}

\section{Experimental details}
\label{apx:details}

In this section, we present the experiment details for the results obtained in the main paper.

\subsection{GMC model training}
We follow the specification in the original GMC paper \cite{poklukar2022geometric} and select a single model checkpoint for all our experiments by early stopping on the GMC validation loss. We employ a Graph Convolutional Network (GCN) for the structure encoder and a simple Multilayer Perceptron (MLP) for the cell morphology inputs. MLPs are also used for the projector architecture for all modalities. See \Cref{tab:gmc_params} for a full breakdown of the hyper-parameters we used.

\begin{table}[!htb]
\centering
\begin{tabular}{ll}
\toprule
\textbf{Parameter} & \textbf{Value} \\
\midrule
Batch size $\beta$ & 128 \\
Number of epochs & 200 \\
Optimizer & Adam \\
Learning rate & $1 \times 2e^{-6}$ \\
Non-Linearity & ReLU \\
Temperature $\tau$ & 0.4 \\
Intermediate Dim. Size $d$ & 1024 \\
Latent Dim. Size $s$ & 1024 \\
\bottomrule \\
\end{tabular}
\caption{Hyperparameters of the Geometric Multimodal Contrastive proxy model.}
\label{tab:gmc_params}
\end{table}

\subsection{Fragment-based molecule generation}

\begin{table}[!htb]
\centering
\begin{tabular}{ll}
\toprule
\textbf{Parameter} & \textbf{Value} \\
\midrule
Batch size & 64 \\
Number of steps & 10,000 \\
Optimizer & Adam \\
Number of Layers & 4 \\
Hidden Dim. Size & 128 \\
Number of Heads & 2 \\
Positional Embeddings & Rotary \\
Reward scaling $\beta$ in $R^\beta$ & 64 \\
Learning rate & $1 \times 10^{-4}$ \\
$Z$ Learning rate & $1 \times 10^{-3}$ \\
\bottomrule \\
\end{tabular}
\caption{Hyperparameters of the Graph Attention Transformer used across all models in fragment-based molecule generation.}
\label{tab:merged_common_gat}
\end{table}

In this section, we provide details on the molecule generation experiments and the hyperparameters we used for the methods presented in the paper. In our experimental setup, we follow the same environment specifications and implementations provided in \cite{malkin2022gflownets} with the exception of a different proxy model (GMC) and reward function. The architecture of the GFlowNet, SAC and SQL models is based on a graph attention transformer \cite{velivckovic2017graph} whose specification is detailed in \Cref{tab:merged_common_gat}. For SAC, we use a fixed $\alpha$ value of 0.2 chosen from $\{0.1, 0.2, 0.5\}$. For SQL, we use a fixed $\alpha$ value of 0.1. All methods use discount factor $\gamma$ value of 1.0.

\subsection{Oracle training}
\label{apx:oracle}
We trained an MLP using molecular fingerprints to predict active biological assays (in a multi-label classification setting) for the 8 targets highlighted on the left of \Cref{fig:targets}. For the input, we use Morgan fingerprints with radius 3 and dimensionality 2048. The MLP has two 64 dimensional hidden layers and uses ReLU activation. We trained the model with a learning rate of $1e^{-4}$ with Adam optimizer for 200 epochs. Model selection was performed based on average precision score on the validation set.

\subsection{Compute Resources}
\label{sec:compute_resources}

All of our experiments were conducted using A100 and V100 GPUs. For the fragment-based generation experiment, we used a single worker and it ran in less than 4 hours. For GMC model training, the runs took a little less than 3 hours. For the oracle training, both MLPs took around an hour to complete training.

\end{document}